\documentclass{article}

\usepackage{microtype}
\usepackage{booktabs} 

\usepackage{hyperref}

\usepackage[accepted]{icml2018}

\usepackage{graphicx}
\usepackage{amsmath}
\usepackage{amssymb}
\usepackage{verbatim}
\usepackage{xcolor}
\usepackage{tikz}
\usepackage{url}
\usepackage{float}
\usepackage{subfig}
\usepackage{xstring}
\usepackage{tikz}
\usetikzlibrary{automata,calc,backgrounds,arrows,positioning, shapes.misc,quotes,arrows.meta}
\usepackage{color}
\usepackage{pgfplots}
\usepackage{mathtools}
\usepackage{array}
\usepackage{multirow}

\newcommand{\x}{{\pmb{x}}}

\newcommand{\z}{{\pmb{z}}}

\newcommand{\h}{{\pmb{h}}}
\renewcommand{\a}{{\pmb{a}}}

\newcommand{\g}{{\pmb{g}}}

\newcommand{\W}{{\pmb{W}}}

\newcommand{\bzeta}{{\pmb{\zeta}}}
\newcommand{\bmu}{{\pmb{\mu}}}
\newcommand{\bnu}{{\pmb{\nu}}}

\newcommand{\E}{{\mathbb{E}}}
\newcommand{\uu}{$\uparrow$}
\newcommand{\dd}{$\downarrow$}
\def\KL{\text{KL}}
\def\H{\text{H}}

\tikzset{
  annotated cuboid/.pic={
    \tikzset{%
      every edge quotes/.append style={midway, auto},
      /cuboid/.cd,
      #1
    }
    \draw [-, every edge/.append style={pic actions, densely dashed, opacity=.5}, pic actions]
    (0,0,0) coordinate (o) -- ++(-\cubescale*\cubex,0,0) coordinate (a) -- ++(0,-\cubescale*\cubey,0) coordinate (b) edge coordinate [pos=1] (g) ++(0,0,-\cubescale*\cubez)  -- ++(\cubescale*\cubex,0,0) coordinate (c) -- cycle
    (o) -- ++(0,0,-\cubescale*\cubez) coordinate (d) -- ++(0,-\cubescale*\cubey,0) coordinate (e) edge (g) -- (c) -- cycle
    (o) -- (a) -- ++(0,0,-\cubescale*\cubez) coordinate (f) edge (g) -- (d) -- cycle;
  },
  /cuboid/.search also={/tikz},
  /cuboid/.cd,
  width/.store in=\cubex,
  height/.store in=\cubey,
  depth/.store in=\cubez,
  units/.store in=\cubeunits,
  scale/.store in=\cubescale,
  width=10,
  height=10,
  depth=10,
  units=cm,
  scale=1.,
}

\icmltitlerunning{DVAE++: Discrete Variational Autoencoders with Overlapping Transformations}

\begin{document}

\twocolumn[
\icmltitle{DVAE++: Discrete Variational Autoencoders with \\ Overlapping Transformations}




\begin{icmlauthorlist}
\icmlauthor{Arash Vahdat}{quad}
\icmlauthor{William G. Macready}{quad}
\icmlauthor{Zhengbing Bian}{quad}
\icmlauthor{Amir Khoshaman}{quad}
\icmlauthor{Evgeny Andriyash}{quad}
\end{icmlauthorlist}

\icmlaffiliation{quad}{Quadrant.ai, D-Wave Systems Inc., Burnaby, BC, Canada}

\icmlcorrespondingauthor{Arash Vahdat}{arash@quadrant.ai}

\icmlkeywords{Machine Learning, ICML, Generative Learning, Discrete Variational Autoencoders}

\vskip 0.3in
]



\printAffiliationsAndNotice{}  

\begin{abstract}
Training of discrete latent variable models remains challenging
because passing gradient information through discrete units is difficult.
We propose a new class of smoothing transformations based on a mixture of two overlapping
distributions, and show that the proposed transformation can be used for training binary latent models with either directed or undirected priors. 
We derive a new variational bound to efficiently train with Boltzmann machine priors.
Using this bound, we develop DVAE++, a generative model with a global discrete prior and a hierarchy of convolutional continuous variables.
Experiments on several benchmarks show that overlapping transformations outperform other recent continuous relaxations of discrete latent variables including Gumbel-Softmax \cite{maddison2016concrete, jang2016categorical}, and discrete variational autoencoders \cite{rolfe2016discrete}.
\end{abstract}

\vspace{-0.5cm}
\section{Introduction}
Recent years have seen rapid progress in generative modeling made possible by advances
in deep learning and stochastic variational inference. The reparameterization trick \cite{kingma2014vae, rezende2014stochastic} 
has made stochastic variational inference efficient by providing lower-variance gradient estimates. However, reparameterization, as originally 
proposed, does not easily extend to semi-supervised learning, binary latent attribute models, topic modeling, variational memory addressing, hard attention models, or clustering, which require discrete latent-variables.

Continuous relaxations have been proposed for accommodating discrete variables in variational inference 
\cite{maddison2016concrete, jang2016categorical, rolfe2016discrete}. 
The Gumbel-Softmax technique \cite{maddison2016concrete, jang2016categorical}
defines a temperature-based continuous distribution that in the zero-temperature limit converges to a discrete distribution. However, it is limited to categorical distributions and 
does not scale to multivariate models such as Boltzmann machines (BM). The approach presented in 
\cite{rolfe2016discrete} can train models with BM priors but requires careful handling of the gradients during training.

We propose a new class of smoothing transformations for relaxing binary latent variables. The method relies on two 
distributions with overlapping support that in the zero temperature limit converge to a Bernoulli distribution.
We present two variants of smoothing transformations using a mixture of exponential and a mixture of logistic distributions.

We demonstrate that overlapping transformations can be used to train discrete directed
latent models as in \cite{maddison2016concrete, jang2016categorical}, \textit{and} models with BMs in their prior as in \cite{rolfe2016discrete}.
In the case of BM priors, we show that the Kullback-Leibler (KL) contribution to the variational bound can be approximated using an analytic expression that can be optimized using automatic differentiation without requiring the special treatment of gradients in \cite{rolfe2016discrete}.

Using this analytic bound, we develop a new variational autoencoder (VAE) architecture called DVAE++, 
which uses a BM prior to model discontinuous latent factors such as object categories or scene configuration in images. 
DVAE++ is inspired by \cite{rolfe2016discrete} and includes continuous local latent variables to model locally 
smooth features in the data. DVAE++ achieves 
comparable results to the state-of-the-art techniques on several datasets 
and captures semantically meaningful discrete aspects of the data. 
We show that even when all continuous latent variables are removed, DVAE++ still attains near state-of-the-art generative 
likelihoods.

\subsection{Related Work}
Training of models with discrete latent variables $\z$ requires low-variance estimates of gradients of the form $\nabla_\phi \mathbb{E}_{q_{\phi}(\z)}[f(\z)]$. Only when $\z$ has a modest number of configurations 
(as in semi-supervised learning \cite{kingma2014semi} or semi-supervised generation \cite{maaloe2017semi}) 
can the gradient of the expectation be decomposed into a summation over configurations.

The REINFORCE technique \cite{williams1992simple} is a more scalable method that migrates the gradient inside the expectation: $\nabla_{\phi} \mathbb{E}_{q_\phi(\z)}f(\z) = \mathbb{E}_{q_{\phi}(\z)} [f(\z) \nabla_{\phi} \log{q_{\phi}(\z)}].$
Although the REINFORCE estimate is unbiased, it suffers from high variance and carefully designed ``control variates" are required to make it practical. 
Several works use this technique and differ in their choices of the control variates. NVIL \cite{mnih2014neural} uses a running average of 
the function, $f(\z)$, and an input-dependent \emph{baseline}. 
VIMCO \cite{mnih2016variational} is a multi-sample version of NVIL that has baselines tailored for each sample based on all the other samples. MuProp \cite{gu2015muprop} and DARN \cite{gregorICML14} are two other 
REINFORCE-based methods (with non-zero biases) that use a Taylor expansion of the function $f(\z)$ to create control variates.

To address the high variance of REINFORCE, other work strives to make discrete variables compatible with the reparametrization technique. A primitive form arises from estimating the discrete variables by a continuous function during back-propagation. For instance, in the case of Bernoulli distribution, the latent variables can be approximated by their mean value. This approach is called the \textit{straight-through (ST) estimator} \cite{bengio2013estimating}.  Another way to make discrete variables compatible with the reparametrization is to relax them into a continuous distribution. Concrete \cite{maddison2016concrete} or Gumbel-Softmax \cite{jang2016categorical} adopt this strategy by adding Gumbel noise to the logits of a softmax function with a temperature hyperparameter. A slope-annealed version of the ST estimator is proposed by \cite{chung2016hierarchical} and is equivalent to the Gumbel-Softmax approach for binary variables. REBAR \cite{tucker2017rebar} is a recent method that blends REINFORCE with Concrete to synthesize control variates. \cite{rolfe2016discrete} pairs discrete variables with auxiliary continuous variables and marginalizes out the discrete variables. 

Both overlapping transformations and Gumbel-based approaches offer smoothing through non-zero temperature; however, overlapping transformations
offer additional freedom through the choice of the mixture distributions. 

\section{Background}

Let $\x$ represent observed random variables and $\z$ latent variables. The joint distribution 
over these variables is defined by the generative model $p(\x, \z) = p(\z) p(\x|\z)$, where $p(\z)$
is a prior distribution and $p(\x|\z)$ is a probabilistic decoder. Given a dataset $\pmb{X} = \{\x^{(1)}, \dots, \x^{(N)}\}$,
the parameters of the model are trained by maximizing the log-likelihood:
\begin{equation}
 \log p(\pmb{X}) = \sum_{i=1}^N \log p(\x^{(i)}). \nonumber
\end{equation}
Typically, computing $\log p(\x)$ requires an intractable marginalization over the latent variables $\z$.
To address this problem, the VAE~\cite{kingma2014vae}
introduces an inference model or probabilistic encoder $q(\z|\x)$ that infers latent
variables for each observation. In the VAE, instead of the maximizing the marginal log-likelihood, a variational lower bound (ELBO) is maximized:
\begin{equation} \label{eq:elbo}
 \log p(\x) \geq \E_{q(\z|\x)}\bigl[\log p(\x|\z)\bigr] - \KL\bigl(q(\z|\x) || p(\z)\bigr).
\end{equation}
The gradient of this objective is computed for the parameters of both the encoder and decoder 
using the reparameterization trick. With reparametrization, the expectation with respect to $q(\z|\x)$ in Eq.~\eqref{eq:elbo}
is replaced with an expectation with respect to a known optimization-parameter-independent base distribution and a 
differentiable transformation from the base distribution to $q(\z|\x)$. This transformation may be a scale-shift transformation, in the case of Gaussian base distributions, or rely on the inverse cumulative distribution function (CDF) in the general case. Following the law of the unconscious statistician, 
the gradient is then estimated using samples from the base distribution.

Unfortunately, the reparameterization trick cannot be applied directly to the discrete latent variables 
because there is no differentiable transformation that maps a base distribution to a discrete distribution. 
Current remedies address this difficulty using a continuous relaxation of the discrete latent variables \cite{maddison2016concrete, jang2016categorical}. 
The discrete variational autoencoder (DVAE) \cite{rolfe2016discrete} develops a different approach 
which applies the reparameterization trick to a marginal distribution constructed by pairing each discrete variable with an auxiliary continuous random 
variable.

For example, let $z \in \{0,1\}$ represent a binary random variable with the probability mass function $q(z|x)$. 
A smoothing transformation is defined using spike-and-exponential transformation $r(\zeta | z)$, 
where $r(\zeta | z=0)=\delta(\zeta)$ is a Dirac $\delta$ distribution and $r(\zeta | z=1) \propto \exp(\beta \zeta)$ 
is an exponential distribution defined for $\zeta \in [0, 1]$
with inverse temperature $\beta$ that controls the sharpness of the distribution. 
\cite{rolfe2016discrete} notes that the autoencoding term can be defined as:
\begin{equation}
\sum_z q(z|x) \!\! \int \!d\zeta \; r(\zeta | z) \log p(x|\zeta) = \int \!\! d\zeta \;  q(\zeta|x) \log p(x|\zeta), \nonumber
\end{equation}
where the marginal 
\begin{equation}
    q(\zeta|x) = \sum_{z} q(z|x) r(\zeta|z) \label{eq:qMarg}
\end{equation}
is a mixture of two continuous distributions. By factoring the inference model so that $x$ depends on $\zeta$ rather 
than $z$, the discrete variables can be explicitly eliminated from the ELBO and the reparameterization trick applied.

The smoothing transformations in \cite{rolfe2016discrete} are limited to spike-and-X type of transformations (e.g., spike-and-exp and spike-and-Gaussian) where $r(\zeta | z=0)$ is assumed to be a Dirac $\delta$ distribution. This property is required for computing the gradient of the KL term in the 
variational lower bound.

\section{Overlapping Transformations} \label{sec:transform}
\begin{figure}
  \centering
    \subfloat[smoothing transformations]{\includegraphics[scale=0.25, trim={0.4cm 0 0.4cm 0},clip]{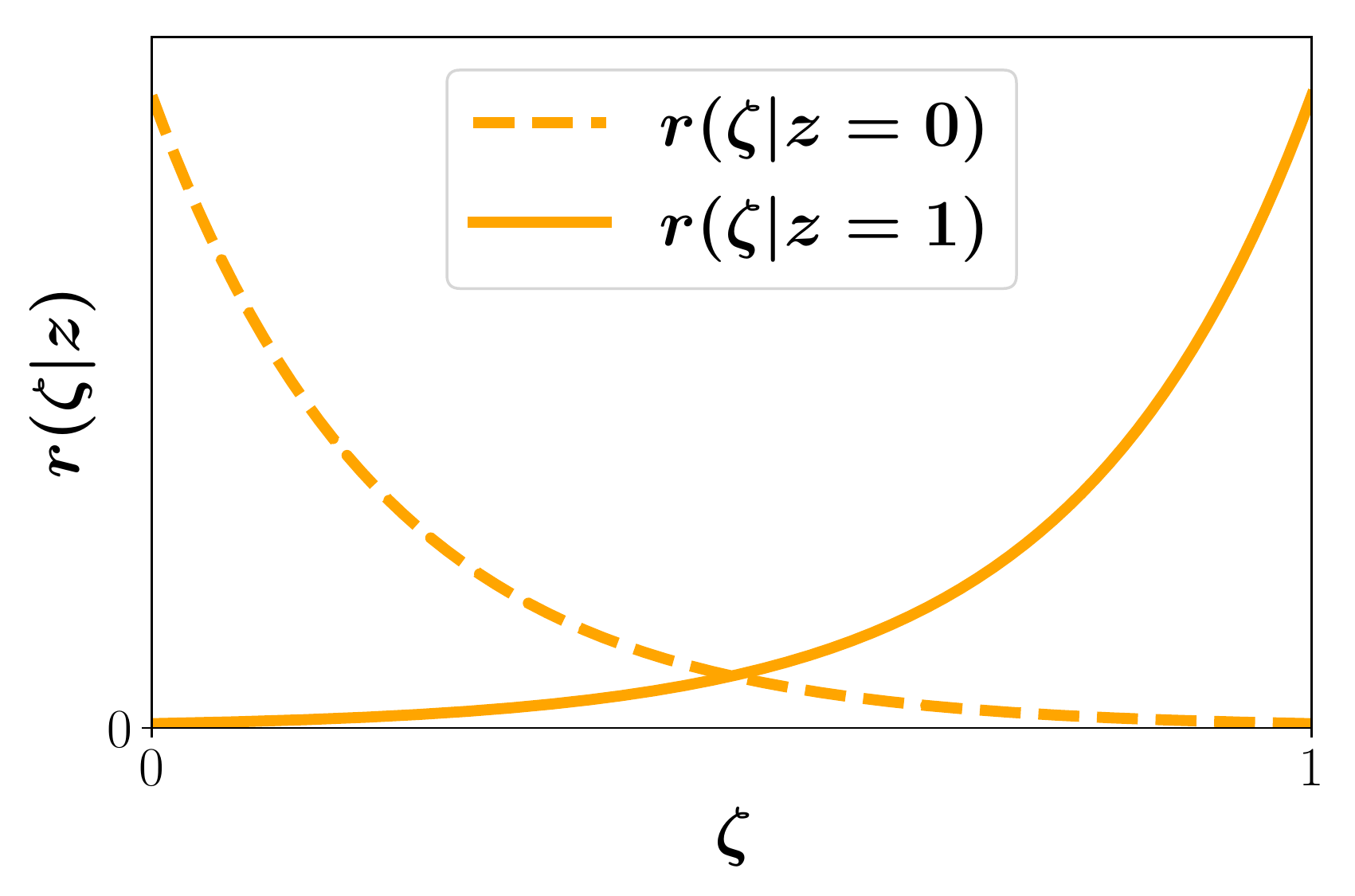}}
    \subfloat[inverse CDF]{\includegraphics[scale=0.25, trim={0.4cm 0 0.4cm 0},clip]{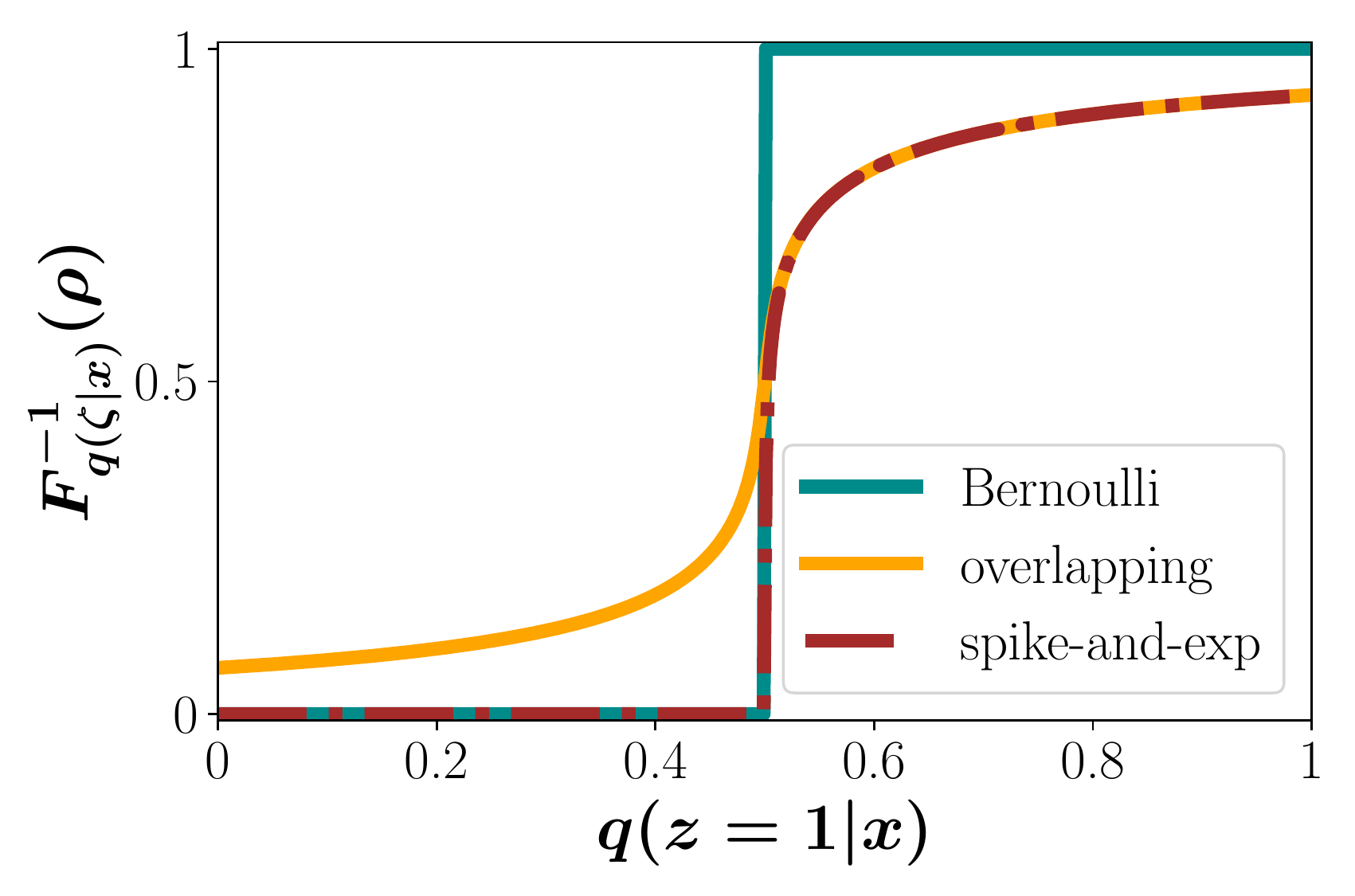}}
    \caption{a) Smoothing transformations using exponential distributions. b) Inverse CDF as a function of $q(z=1|x)$ for $\rho=0.5$ in comparison to the spike-and-exp smoothing \cite{rolfe2016discrete}. The inverse CDF resulting from the mixture of exponential distributions approximates the
    step function that samples from the Bernoulli distribution.
    }
     \label{fig:exp_smooth}
\end{figure}

A symmetric smoothing transformation of binary variables can also be defined using two exponential distributions:
\begin{align*}
r(\zeta | z = 0) = \frac{e^{- \beta \zeta}}{Z_\beta}       \quad \text{and} \quad r(\zeta | z = 1) = \frac{e^{\beta (\zeta - 1)}}{Z_\beta},
\end{align*}
for $\zeta\in [0,1]$, where $Z_\beta = (1 - e^{-\beta})/\beta$. These conditionals, visualized in Fig.~\ref{fig:exp_smooth}(a),
define the mixture distribution $q(\zeta|x)$ of Eq.~\eqref{eq:qMarg}.
The scalar $\beta$ acts as an inverse temperature as in the Gumbel softmax relaxation, and as $\beta \rightarrow \infty$, $q(\zeta|x)$ approaches $q(z=0|x)\delta(\zeta) + q(z=1|x)\delta(\zeta-1)$.

Application of the reparameterization trick for $q(\zeta|x)$ requires the inverse CDF of $q(\zeta|x)$. In Appendix~\ref{app:overlap} of the supplementary material, we show that the inverse CDF is
\begin{equation}
F^{-1}_{q(\zeta|x)}(\rho)= -\frac{1}{\beta} \log{\frac{-b + \sqrt{b^2 - 4c}}{2}} \label{eq:inv_mix}
\end{equation}
where $b =  [\rho + e^{-\beta}(q - \rho)]/(1 - q) -1$ and $c = -[q e^{-\beta}]/(1-q)$. 
Eq.~\eqref{eq:inv_mix} is a differentiable function that
converts a sample $\rho$ from the uniform distribution $\mathcal{U}(0,1)$ to a sample
from $q(\zeta|x)$. As shown in Fig.~\ref{fig:exp_smooth}(b) the inverse CDF approaches a step function as $\beta\rightarrow\infty$. However, to benefit from gradient information during training, $\beta$ is set to a finite value. Appendix~\ref{app:invCDFViz} provides further visualizations comparing overlapping transformations to Concrete smoothing~\cite{maddison2016concrete, jang2016categorical}.

The overlapping exponential distributions defined here can be generalized to any pair of smooth 
distributions converging to $\delta(\zeta)$ and $\delta(\zeta-1)$. In Appendix~\ref{app:otherOverlap}, we provide analogous results for logistic smoothing distributions.

Next, we apply overlapping transformations to the training of generative models with discrete latent variables. We consider both directed and 
undirected latent variable priors.

\begin{figure}
 \centering
   \subfloat[]{
  \begin{tikzpicture}[->,>=stealth',shorten >=1pt,auto,node distance=1.5cm, thick, scale=0.7]
      \tikzstyle{every state}=[fill=white,draw=black,text=black, transform shape]
      \node[state] 			 (z1)                   {$\z_1$};
      \node[state] 			 (z2)   [below of=z1]   {$\z_2$};
      \node[state] 			 (x)    [below of=z2]   {$\x$};
      \path (z1)        edge [bend left] (x);
      \path (z2)        edge             (x);
      \path (z1)        edge             (z2);
  \end{tikzpicture}  
  }\hspace{-1mm}%
   \subfloat[]{
 \begin{tikzpicture}[->,>=stealth',shorten >=1pt,auto,node distance=1.5cm, thick, scale=0.7]
      \tikzstyle{every state}=[fill=white,draw=black,text=black, transform shape]
      \node[state] 			   (x)                    {$\x$};
      \node[state] 			   (z1)   [below of=x]    {$\z_1$};
      \node[state] 			   (z2)   [below of=z1]   {$\z_2$};
      \path (x)         edge               (z1);
      \path (z1)        edge               (z2);
      \path (x)         edge [bend left]   (z2);
  \end{tikzpicture}  
  }\hspace{-1mm}%
  \subfloat[]{
  \begin{tikzpicture}[->,>=stealth',shorten >=1pt,auto,node distance=1.5cm, thick, scale=0.7]
      \tikzstyle{every state}=[fill=white,draw=black,text=black, transform shape]
      \node[state] 		(z1)                   {$\z_1$};
      \node[state] 		(zeta1)   [below of=z1]   {$\bzeta_1$};
      \node[state] 		(z2)      [right= 0.12cm of z1] {$\z_2$};
      \node[state] 		(zeta2)   [below of=z2]   {$\bzeta_2$};
      \node[state] 			 (x)    [below of=zeta2]   {$\x$};
      \path (z1)        edge             (zeta1);
      \path (z2)        edge             (zeta2);
      \path (zeta1)        edge             (x);
      \path (zeta2)        edge             (x);
      \path (zeta1)    edge  (z2) ;
  \end{tikzpicture}  
  }\hspace{-1mm}%
  \subfloat[]{
 \begin{tikzpicture}[->,>=stealth',shorten >=1pt,auto,node distance=1.5cm, thick, scale=0.7]
      \tikzstyle{every state}=[fill=white,draw=black,text=black, transform shape]
      \node[state] 	(x)                       {$\x$};
      \node[state] 	(z1)     [below of=x]    {$\z_1$};
      \node[state] 	(zeta1)  [below of=z1]   {$\bzeta_1$};
      \node[state] 	(z2)     [right= 0.12cm of z1]    {$\z_2$};
      \node[state] 	(zeta2)  [below of=z2]   {$\bzeta_2$};
      \path (x)         edge               (z1);
      \path (x)         edge               (z2);
      \path (z1)        edge               (zeta1);
      \path (zeta1)    edge  (z2) ;
      \path (z2)        edge               (zeta2);
  \end{tikzpicture}  
  }\hspace{-1mm}%
  \subfloat[]{
  \begin{tikzpicture}[->,>=stealth',shorten >=1pt,auto,node distance=1.5cm, thick, scale=0.7]
      \tikzstyle{every state}=[fill=white,draw=black,text=black, transform shape]
      \node[state] 			 (z1)                   {$\bzeta_1$};
      \node[state] 			 (z2)   [below of=z1]   {$\bzeta_2$};
      \node[state] 			 (x)    [below of=z2]   {$\x$};
      \path (z1)        edge [bend left] (x);
      \path (z2)        edge             (x);
      \path (z1)        edge             (z2);
  \end{tikzpicture}  
  }\hspace{-1mm}%
   \subfloat[]{
 \begin{tikzpicture}[->,>=stealth',shorten >=1pt,auto,node distance=1.5cm, thick, scale=0.7]
      \tikzstyle{every state}=[fill=white,draw=black,text=black, transform shape]
      \node[state] 			   (x)                    {$\x$};
      \node[state] 			   (z1)   [below of=x]    {$\bzeta_1$};
      \node[state] 			   (z2)   [below of=z1]    {$\bzeta_2$};
      \path (x)         edge               (z1);
      \path (z1)        edge               (z2);
      \path (x)         edge [bend left]   (z2);
  \end{tikzpicture}  
  }\hspace{-1mm}%
  \subfloat[]{
  \begin{tikzpicture}[->,>=stealth',shorten >=1pt,auto,node distance=1.5cm, thick, scale=0.7]
      \tikzstyle{every state}=[fill=white,draw=black,text=black, transform shape]
      \node[state] 		(z1)                   {$\z_1$};
      \node[state] 		(z1)                   {$\z_1$};
      \node[state] 		(zeta1)   [below of=z1]   {$\bzeta_1$};
      \node[state] 		(z2)      [right= 0.15cm of z1] {$\z_2$};
      \draw [dashed, rounded corners=.3cm]
        (-0.6,-0.6) rectangle (1.8,0.6);
      \node[state] 		(zeta2)   [below of=z2]   {$\bzeta_2$};
      \node[state] 			 (x)    [below of=zeta2]   {$\x$};
      \path (z1)        edge             (zeta1);
      \path (z2)        edge             (zeta2);
      \path (zeta1)     edge             (x);
      \path (zeta2)     edge             (x);
      \path (z1)        edge [-]           (z2) ;
  \end{tikzpicture}
  }
  \caption{(a) A generative model with binary latent variables $\z_1$ and $\z_2$, and (b) the corresponding inference model. In (c) and (d), the continuous $\zeta$ is introduced and dependencies on $\z$ are transferred to dependencies on $\zeta$. In (e) and (f)
  the binary latent variables $\z$ are marginalized out. (g) A generative model with a Boltzmann machine (dashed) prior.}
  \label{fig:sbn}
\end{figure}
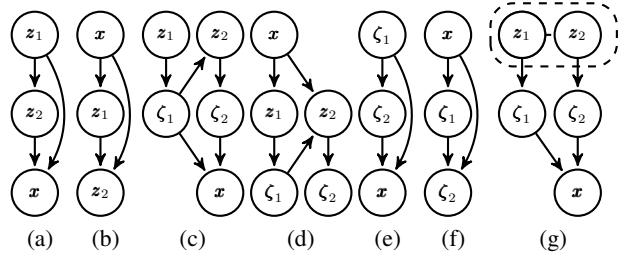

\section{Directed Prior}
The simplest discrete prior is factorial; however, with conditioning, we can build complex dependencies. To simplify presentation, we illustrate
a VAE prior with one or two groups of conditioning variables, but note that the approach straight-forwardly generalizes
to many conditioning groups.

Our approach parallels the method developed in \cite{rolfe2016discrete} for undirected graphical models. 
Consider the generative model in Fig.~\ref{fig:sbn}(a) and its corresponding inference model in Fig.~\ref{fig:sbn}(b). 
To train this model using smoothing transformations, we introduce the continuous $\zeta$ in Figs.~\ref{fig:sbn}(c) and \ref{fig:sbn}(d) in which
dependencies on $z$ are transferred to dependencies on $\zeta$. In this way, binary latent variables influence other variables only through 
their continuous counterparts. In Figs.~\ref{fig:sbn}(e) and \ref{fig:sbn}(f) we show the same model but with $\z$ marginalized out. 
The joint $(\z,\bzeta)$ 
model of Figs.~\ref{fig:sbn}(c) and \ref{fig:sbn}(d) gives rise to a looser ELBO than the marginal $\bzeta$ model of Figs.~\ref{fig:sbn}(e) and \ref{fig:sbn}(f).

\subsection{Joint ELBO}

Assuming that $p(\z_1)$, $p(\z_2|\bzeta_1)$, $q(\z_1|\x)$, $q(\z_2|\x, \bzeta_1)$, $r(\bzeta_1|\z_1)$, and $r(\bzeta_2|\z_2)$ 
are factorial in both the inference and generative models, then $q(\bzeta_1|\x)$ and $q(\bzeta_2|\bzeta_1,\x)$ are also
factorial with $q(\bzeta_{1}|\x) = \prod_i q(\zeta_{1,i}|\x)$ where $q(\zeta_{1,i}|\x)= \sum_{z_{1,i}} r(\zeta_{1,i}|z_{1,i}) q(z_{1,i}|\x)$, 
and $q(\bzeta_2|\bzeta_1,\x) = \prod_i q(\zeta_{2,i}|\bzeta_1,\x)$ where $q(\zeta_{2,i}|\bzeta_1,\x) = 
\sum_{z_{2,i}} r(\zeta_{2,i}|z_{2,i}) q(z_{2,i}|\bzeta_1,\x)$.
In this case, the ELBO for the model in Fig.~\ref{fig:sbn}(c) and \ref{fig:sbn}(d) is
\begin{flalign}
 &\E_{q(\bzeta_1|\x)}\left[ \E_{q(\bzeta_2|\bzeta_1,\x)} \left[\log p(\x|\bzeta_1, \bzeta_2) \right]\right] - \KL(q(\z_1|\x) || p(\z_1)) \nonumber \\
 & \quad - \E_{q(\bzeta_1|\x)} \left[ \KL(q(\z_2|\x, \bzeta_1) || p(\z_2|\bzeta_1)) \right]. \label{eq:elbo_sbn}
\end{flalign}
The KL terms corresponding to the divergence between factorial Bernoulli distributions have a closed form.
The expectation over $\bzeta_1$ and $\bzeta_2$ is reparameterized using the technique presented in Sec.~\ref{sec:transform}.

\subsection{Marginal ELBO} \label{sec:cont_relax}
The ELBO for the marginal graphical model of Fig.~\ref{fig:sbn}(e) and Fig.~\ref{fig:sbn}(f) is
\begin{flalign}
 &\E_{q(\bzeta_1|\x)}\left[ \E_{q(\bzeta_2|\x, \bzeta_1)} \left[\log p(\x|\bzeta_1, \bzeta_2) \right]\right] - \KL(q(\bzeta_1|\x) || p(\bzeta_1)) \nonumber \\
 & \quad - \E_{q(\bzeta_1|\x)} \left[ \KL(q(\bzeta_2|\x, \bzeta_1) || p(\bzeta_2|\bzeta_1)) \right] \label{eq:elbo_sbn2}
\end{flalign}
with $p(\bzeta_1) = \prod_i p(\zeta_{1, i})$ where $p(\zeta_{1, i}) = \sum_{z_i} r(\zeta_{1,i}|z_{1,i}) p(z_{1,i})$ and $p(\bzeta_2|\bzeta_1)= \prod_i p(\zeta_{2,i}|\bzeta_1)$ where $p(\zeta_{2,i}|\bzeta_1) = \sum_{z_{2,i}} r(\zeta_{2,i}|z_{2,i}) p(z_{2,i}|\bzeta_1)$. 
The KL terms no longer have a closed form but can be estimated with the Monte Carlo method.
In Appendix~\ref{app:elbos}, we show that Eq.~\eqref{eq:elbo_sbn2} provides a tighter bound on $\log p(\x)$ than does Eq.~\eqref{eq:elbo_sbn}.

\section{Boltzmann Machine Prior} 
\cite{rolfe2016discrete} defined an expressive prior over binary latent variables by using a Boltzmann machine.
We build upon that work and present a simpler objective that can still be trained with a low-variance gradient estimate.

To simplify notation, we assume that the prior distribution over the latent binary variables is a restricted Boltzmann machine (RBM), but these results can be extended to general BMs. An RBM defines a probability distribution over binary random variables arranged on a bipartite graph as $p(\z_1, \z_2) = e^{-E(\z_1,\z_2)}/Z$ where $E(\z_1,\z_2) = -\a_1^T \z_1 - \a_2^T\z_2 - \z_1^T \W \z_2$ is an energy function with linear biases $\a_1$ and $\a_2$, and pairwise interactions $\W$. $Z$ is the partition function.

Fig.~\ref{fig:sbn}(g) visualizes a generative model with a BM prior.
As in Figs.~\ref{fig:sbn}(c) and \ref{fig:sbn}(d), conditionals are formed on the auxiliary variables $\bzeta$ instead of the binary variables $\z$. The 
inference model in this case is identical to the model in Fig.~\ref{fig:sbn}(d) and it infers both $\z$ and $\bzeta$ in a hierarchical structure.

The autoencoding contribution to the ELBO with an RBM prior is again the first term in Eq.~\eqref{eq:elbo_sbn} since both models share the same inference model structure. However, computing the KL term with the RBM prior is more challenging. Here, a novel formulation for the KL term is introduced. Our derivation can be used for training discrete variational autoencoders with a BM prior without any manual coding of gradients.

We use $\E_{q(\z, \bzeta|\x)}[f] = \E_{q(\bzeta|\x)}\bigl[\E_{q(\z|\x, \bzeta)}[f]\bigr]$ to compute the KL contribution to the ELBO:
{\small
\begin{align}
 &\KL\bigl(q(\z_1, \z_2, \bzeta_1, \bzeta_2 | \x) \| p(\z_1, \z_2, \bzeta_1, \bzeta_2)\bigr) = \nonumber \\
 & \ \log Z -\H\bigl(q(\z_1|\x)\bigr) - \E_{q(\bzeta_1|\x)} \left[\H\bigl(q(\z_2|\x, \bzeta_1)\bigr)\right] + \label{eq:kl_simple} \\
 & +\E_{q(\bzeta_1|\x)}\!\big[\E_{q(\bzeta_2|\x, \bzeta_1)}\!\big[\underbrace{\E_{q(\z_1|\x, \bzeta_1)}\!\big[
 \E_{q(\z_2|\x, \bzeta_1, \bzeta_2)}\!\big[ E(\z_1, \z_2) \big]\big]}_{\text{cross-entropy}}\big]\big]. \nonumber
\end{align}}
Here, $\H(q)$ is the entropy of the distribution $q$, which has a closed form when $q$ is factorial Bernoulli. The conditionals $q(\z_1|\x, \bzeta_1)$ and $q(\z_2|\x, \bzeta_1, \bzeta_2)$ are both
factorial distributions that have analytic expressions. Denoting 
\begin{align*}
\mu_{1,i}(\x) &\equiv q(z_{1,i}=1|\x),  \\
\nu_{1,i}(\x, \bzeta_1) &\equiv q(z_{1,i}=1|\x, \bzeta_1), \\
\mu_{2,i}(\x, \bzeta_1) &\equiv q(z_{2,i}=1|\x, \bzeta_1), \\
\nu_{2,i}(\x, \bzeta_1, \bzeta_2) &\equiv q(z_{2,i}=1|\x, \bzeta_1, \bzeta_2),
\end{align*}
it is straightforward to show that 
\begin{align*}
\footnotesize
    \nu_{1, i}(\x, \bzeta_1) &= \frac{q(z_{1,i}=1|\x) r(\zeta_{1, i}|z_{1, i}=1)}{\sum_{z_{1,i}}q(z_{1,i}|\x) r(\zeta_{1, i}|z_{1, i})} = \\
    &= \sigma\Bigl(g(\mu_{1,i}(\x)) + \log\bigl[\frac{r(\zeta_{1, i}|z=1)}{r(\zeta_{1, i}|z=0)}\bigr]\Bigr),
\end{align*}
where $\sigma(x)=1/(1+e^{-x})$ is the logistic function, and $g(\mu) \equiv \log \bigl[\mu/\bigl(1 - \mu\bigr)\bigr]$ is the logit function. A similar expression holds for $\bnu_2(\x, \bzeta_1, \bzeta_2)$. The expectation marked as cross-entropy in 
Eq.~\eqref{eq:kl_simple} corresponds to the cross-entropy between a factorial distribution and an unnormalized Boltzmann machine which is
\begin{equation*}
- \a_1^T \bnu_1(\x, \bzeta_1) -  \a_2^T \bnu_2(\x, \bzeta_1, \bzeta_2) - \bnu_1(\x, \bzeta_1)^T \W \bnu_2(\x, \bzeta_1, \bzeta_2).
\end{equation*}
Finally, we use the equalities 
$\E_{q(\bzeta_1|\x)}[\bnu_1(\x, \bzeta_1)] = \bmu_1(\x)$ and
$\E_{q(\bzeta_2|\x, \bzeta_1)}[\bnu_2(\x, \bzeta_1, \bzeta_2)] = \bmu_2(\x, \bzeta_1)$
to simplify the cross-entropy term which defines the KL as
\begin{align}
 &\KL\bigl(q(\z_1, \z_2, \bzeta_1, \bzeta_2 | \x) \| p(\z_1, \z_2, \bzeta_1, \bzeta_2)\bigr) =  \log Z \nonumber \\
 & -\H\bigl(q(\z_1|\x)\bigr) - \E_{q(\bzeta_1|\x)} \left[\H\bigl(q(\z_2|\x, \bzeta_1)\bigr)\right] \nonumber \\
 &- \a_1^T \bmu_1(\x) - \E_{q(\bzeta_1|\x)}\left[ \a_2^T \bmu_2(\x, \bzeta_1) \right] \nonumber \\
 &-\E_{q(\bzeta_1|\x)} \left[ \bnu_1(\x, \bzeta_1)^T \W \bmu_2(\x, \bzeta_1) \right]. \nonumber
\end{align}
All terms contributing to the KL other than $\log Z$ can be computed analytically given samples from the hierarchical encoder.
Expectations with respect to $q(\bzeta_1|\x)$ are reparameterized using the inverse CDF function. Any automatic differentiation (AD) library can then back-propagate gradients through the network. Only $\log Z$ requires special treatment. In Appendix~\ref{app:gradZ},
we show how this term can also be included in the objective function so that its gradient is computed automatically. The ability of AD to calculate gradients stands in contrast to \cite{rolfe2016discrete} where gradients must be manually coded. This pleasing property is a result of $r(\zeta|z)$ 
having the same support for both $z=0$ and $z=1$, and
having a probabilistic $q(z|x, \zeta)$ which is not the case for the spike-and-X transformations of  \cite{rolfe2016discrete}.

\section{DVAE++}

In previous sections, we have illustrated with simple examples how overlapping transformations can be used to train discrete latent variable models with either directed or undirected priors. Here, we develop a network architecture (DVAE++) that improves
upon convolutional VAEs for generative image modeling.

DVAE++ features both global discrete latent variables (to capture global properties such as scene or object type) and local continuous latent variables (to capture local properties 
such as object pose, orientation, or style). Both generative and inference networks rely on an autoregressive structure defined over groups of latent and observed variables. As we are modeling images, conditional dependencies between groups of variables are captured with convolutional neural networks. DVAE++ is similar to
the convolutional VAEs used in \cite{kingma2016improved, chen2016variational}, but does not use normalizing flows.

\subsection{Graphical Model}

\begin{figure*}
 \centering
   \subfloat[generative model]{
  \begin{tikzpicture}[->,>=stealth',shorten >=1pt,auto,node distance=1.5cm, thick, scale=0.75]
       \tikzstyle{every state}=[fill=white,draw=black,text=black, transform shape]
       \pic [fill=white, draw=black] at (0,0) {annotated cuboid={width=0.5, height=1.5, depth=1.5}};
       \pic [fill=white, draw=black] at (2,0) {annotated cuboid={width=0.5, height=1.5, depth=1.5}};
       \pic [fill=white, draw=black] at (4,0) {annotated cuboid={width=0.5, height=1.5, depth=1.5}};
       \node[] at (-0.25,-0.8) {$\h_1$};
       \node[] at (1.75,-0.8) {$\h_2$};
       \node[] at (3.75,-0.8) {$\h_3$};
       
       \coordinate (a) at (0, -1.0, -1);
       \coordinate (b) at (1.0, -1.0, -1);
       \coordinate (c) at (2.0, -1.0, -1);
       \coordinate (d) at (3.0, -1.0, -1);
       \path[->, thick] (a)  edge (b);
       \path[->, thick] (c)  edge (d);
       \path[->, thick] (a)  edge [bend right=50] (d);
       \node at (6.0,-2.5) [shape=rectangle,draw, minimum size=0.7cm] (x) {$\x$};
       \coordinate (a1) at (-0.25, -1.8, 0.);
       \coordinate (b1) at (1.75, -1.8, 0.);
       \coordinate (c1) at (3.75, -1.8, 0.);
       \draw (a1) to [out=-90,in=180] (2.0, -2.5) to (x) ;
       \draw (b1) to [out=-90,in=180] (3.0, -2.5) to [out=0,in=180] (x) ;
       \draw (c1) to [out=-90,in=180] (5.0, -2.5) to [out=0,in=180] (x) ;
       \node at (1.0, 3.3) [shape=rectangle,draw, minimum size=0.6cm] (z01) {$\ \quad \z_{1}\quad \ $};
       \node at (1.0, 2) [shape=rectangle,draw, minimum size=0.6cm] (zeta01) {$\ \quad \bzeta_{1}\quad \ $};
       \node at (4.0, 3.3) [shape=rectangle,draw, minimum size=0.6cm] (z02) {$\ \quad \z_{2}\quad \ $};
       \node at (4.0, 2) [shape=rectangle,draw, minimum size=0.6cm] (zeta02) {$\ \quad \bzeta_{2}\quad \ $};
       \draw [dashed, rounded corners=.2cm] (-0.1,2.8) rectangle (5.1,3.8);
       \path[-, thick] (z01)  edge [] (z02);
       \coordinate (a2) at (-0.35, 0.2, -1.5);
       \coordinate (a3) at (-0.25, 0.2, -1.5);
       \coordinate (b2) at (1.75, 0.2, -1.5);
       \coordinate (c2) at (3.65, 0.2, -1.5);
       \path[->, thick] (zeta01)  edge [] (a2);
       \path[->, thick] (zeta01)  edge [] (b2);
       \path[->, thick] (zeta01)  edge [] (c2);
       \path[->, thick] (zeta02)  edge [] (a3);
       \path[->, thick] (zeta02)  edge [] (b2);
       \path[->, thick] (zeta02)  edge [] (c2);
       \path[->, thick] (z01)  edge [] (zeta01);
       \path[->, thick] (z02)  edge [] (zeta02);
       \draw (zeta01) to [out=180,in=180] (-0.0, -2.5) to [out=0,in=180] (x) ;
       \draw (zeta02) to [out=0,in=90] (6.0, 1.0) to [out=-90,in=90] (x) ;
  \end{tikzpicture}  
  } \hspace{1cm}
   \subfloat[inference model]{
  \begin{tikzpicture}[->,>=stealth',shorten >=1pt,auto,node distance=1.5cm, thick, scale=0.75]
      \tikzstyle{every state}=[fill=white,draw=black,text=black, transform shape]
       \pic [fill=white, draw=black] at (0,0) {annotated cuboid={width=0.5, height=1.5, depth=1.5}};
       \pic [fill=white, draw=black] at (2,0) {annotated cuboid={width=0.5, height=1.5, depth=1.5}};
       \pic [fill=white, draw=black] at (4,0) {annotated cuboid={width=0.5, height=1.5, depth=1.5}};
       \node[] at (-0.25,-0.8) {$\h_1$};
       \node[] at (1.75,-0.8) {$\h_2$};
       \node[] at (3.75,-0.8) {$\h_3$};
       
       \coordinate (a) at (0, -1.0, -1);
       \coordinate (b) at (1.0, -1.0, -1);
       \coordinate (c) at (2.0, -1.0, -1);
       \coordinate (d) at (3.0, -1.0, -1);
       \path[->, thick] (a)  edge (b);
       \path[->, thick] (c)  edge (d);
       \path[->, thick] (a)  edge [bend right=50] (d);
       \node at (-2.0,-2.5) [shape=rectangle,draw, minimum size=0.7cm] (x) {$\x$};
       \coordinate (a1) at (-0.25, -1.7, 0.);
       \coordinate (b1) at (1.75, -1.7, 0.);
       \coordinate (c1) at (3.75, -1.7, 0.);
       \draw (x) to (-1.0, -2.5) to [out=0,in=-110] (a1) ;
       \draw (x) to ( 1.0, -2.5) to [out=0,in=-110] (b1) ;
       \draw (x) to ( 3.0, -2.5) to [out=0,in=-110] (c1) ;
       \node at (0.25, 2) [shape=rectangle,draw, minimum size=0.6cm] (zeta01) {$\ \quad \bzeta_{1}\quad \ $};
       \node at (3.25, 2) [shape=rectangle,draw, minimum size=0.6cm] (zeta02) {$\ \quad \bzeta_{2}\quad \ $};
       \node at (0.25, 3.3) [shape=rectangle,draw, minimum size=0.6cm] (z01) {$\ \quad \z_{1}\quad \ $};
       \node at (3.25, 3.3) [shape=rectangle,draw, minimum size=0.6cm] (z02) {$\ \quad \z_{2}\quad \ $};
       \path[->, thick] (z01)  edge [] (zeta01);
       \path[->, thick] (z02)  edge [] (zeta02);
       \draw (zeta01)  to [out=10,in=190] (z02) ;
       \coordinate (a2) at (-0.25, 0.2, -1.5);
       \coordinate (b2) at (1.75, 0.2, -1.5);
       \coordinate (c2) at (3.65, 0.2, -1.5);
       \coordinate (c3) at (3.85, 0.2, -1.5);
       \path[->, thick] (zeta01)  edge [] (a2);
       \path[->, thick] (zeta01)  edge [] (b2);
       \path[->, thick] (zeta01)  edge [] (c2);
       \path[->, thick] (zeta02)  edge [] (a2);
       \path[->, thick] (zeta02)  edge [] (b2);
       \path[->, thick] (zeta02)  edge [] (c3);
       \draw (x) to (-2.0, 2.0) to [out=90,in=-180] (z01) ;
       \draw (x) to (-2.0, 2.0) to [out=90,in=-215] (z02) ;
  \end{tikzpicture}  
  } 
  \caption{a) In the generative model, binary global latent variables $\z_1$ and $\z_2$ are modeled by an RBM (dashed) and a series of local continuous variables are generated in an autoregressive structure using residual networks. b) After forming distributions over the global variables, the inference model defines the conditional on the local latent variables similarly using residual networks.}
  \label{fig:pgm}
\end{figure*}
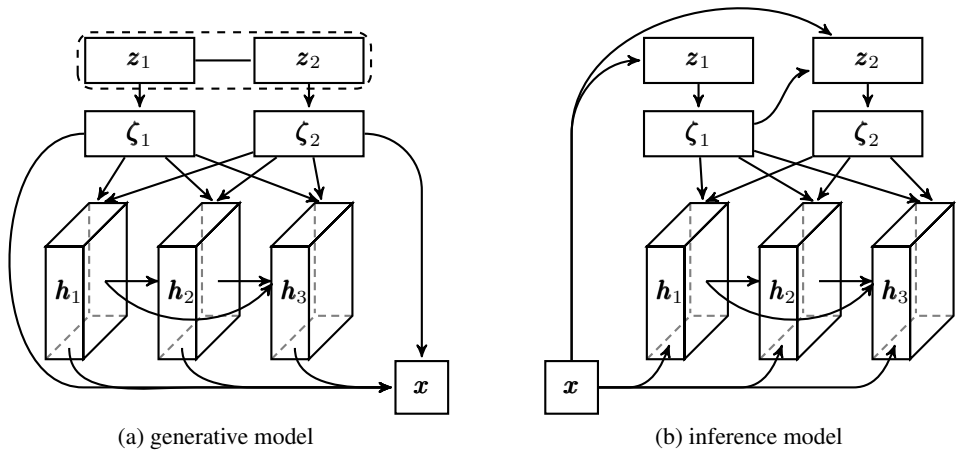

The DVAE++ graphical model is visualized in Fig.~\ref{fig:pgm}. Global and local variables are indicated by $\z$ and $\h$ respectively. Subscripts indicate different groups of random variables. The conditional distribution of each group is factorial --except for $\z_1$ and $\z_2$ in the prior, which is modeled with an RBM. Global latent variables are represented with boxes and local variables are represented with 3D volumes as they are convolutional. 

Groups of local continuous variables are factorial (independent). This assumption limits the ability of the model to capture correlations at different spatial locations and different depths. While the autoregressive structure mitigates this defect, we rely mainly on the discrete global latent variables to capture long-range dependencies. The discrete nature of the global RBM prior allows DVAE++ to capture richly-correlated discontinuous hidden factors that influence data generation. 

Fig.~\ref{fig:pgm}(a) defines the generative model as
\begin{align*}
 p(\z, \bzeta, \h, \x) =& \, p(\z) \prod_i r(\zeta_{1,i}|z_{1,i}) r(\zeta_{2,i}|z_{2,i}) \times \\ & \prod_j p(\h_j | \h_{<j}, \bzeta)p(\x|\bzeta, \h)
\end{align*}
where $p(\z)$ is an RBM, $\bzeta = [\bzeta_1,\bzeta_2]$, and $r$ is the smoothing transformation that is applied 
elementwise to $\z$.
The conditional $p(\h_j | \h_{<j}, \bzeta)$ is defined over the $j^{th}$ local variable
group using a factorial normal distribution. Inspired by \cite{reedParallel2017, denton2015deep}, the conditional on the data variable
$p(\x|\bzeta, \h)$ is decomposed into several factors defined on different scales of $\x$:
\begin{equation*}
 p(\x|\bzeta, \h) = p(\x_0|\bzeta, \h) \prod_{i} p(\x_i|\bzeta, \h, \x_{<i})
\end{equation*}
Here, $\x_0$ is of size $4\times4$ and it represents downsampled $\x$ in the lowest scale. 
Conditioned on $\x_0$, we generate $\x_1$ in the next scale, which is of the size $8\times8$.
This process is continued until the full-scale image is generated (see Appendix~\ref{app:cond_decod} for more details). Here,
each conditional is represented using a factorial distribution. For binary images, a
factorial Bernoulli distribution is used; for colored images a factorial mixture of
discretized logistic distributions is used \cite{salimans2017pixelcnn++}.

The inference model of Fig.~\ref{fig:pgm}(b) conditions over latent variables in a similar order as the generative model:
\begin{align*}
 q(\z, \bzeta, \h | \x) =& \ q(\z_1 | \x) \prod_i r(\zeta_{1,i}|z_{1,i}) \times \\
 & \ q(\z_2|\x, \bzeta_1) \prod_k r(\zeta_{2,k}|z_{2,k}) \prod_j q(\h_j | \bzeta, \h_{<j}).
\end{align*}
The conditionals $q(\z_1 | \x)$ and $q(\z_2|\x, \bzeta_1)$ are each modeled with a factorial Bernoulli distribution, and $q(\h_j | \bzeta, \h_{<j})$ represents the conditional on the $j^{th}$ group of local variables.

DVAE++ is related to VAEs with mixture priors \cite{makhzani2015adversarial, tomczak2017vae}. 
The discrete variables $\z_1$ and $\z_2$ take exponentially many joint configurations where each configuration corresponds to a mixture component. These components are mixed by $p(\z_1, \z_2)$ in the generative model. During training, the inference model maps each data
point to a small subset of all the possible mixture components. Thus, the discrete prior learns to suppress the probability of configurations
that are not used by the inference model. Training results in a multimodal $p(\z_1, \z_2)$ that assigns similar images to a common discrete mode.

\subsection{Neural Network Architecture} \label{app:nnArch}
We use a novel neural network architecture to realize the conditional probabilities within the graphical model Fig.~\ref{fig:pgm}. The network uses residual connections \cite{he2016deep} with squeeze-and-excitation (SE) blocks \cite{hu2017squeeze} 
that have shown state-of-the-art image classification performance. Our architecture is explained fully in Appendix~\ref{app:impDetails}, and here we sketch the main components. We refer to a SE-ResNet block as a residual block, and the network is created by combining either residual blocks, fully-connected layers, or convolutional layers.

The encoder uses a series of downsampling residual blocks to extract convolutional features from an input image. This residual network is considered as a pre-processing step that extracts convolutional feature maps at different scales. 
The output of this network at the highest level is fed to
fully-connected networks that define $q(\z_i | \x, \bzeta_{<i})$ successively for all the global latent variables. 
The feature maps at an intermediate scale are fed to another set of residual networks that define $q(\h_j | \x, \bzeta, \h_{<j})$ successively for all the local latent variables.

The decoder uses an upsampling network to scale-up the global latent variables to the intermediate scale. Then, the output of this network is fed to a set of residual networks that define $p(\h_j | \bzeta, \h_{<j})$ one at a time at the same scale. Finally, another set of residual networks progressively scales the samples from the latent variables up to the data space. In the data space, a distribution on the smallest scale $\x_0$ is formed using a residual network. Given samples at this scale, the distribution at the
next scale is formed using another upsampling residual network. This process is repeated until the image is generated at full scale. 

With many layers of latent variables, the VAE objective often turns off many of the latent variables by matching 
their distribution in the inference model to the prior. The latent units are usually removed differentially across different groups. 
Appendix~\ref{app:KLBalance} presents a technique that enables efficient use of latent variables across all groups.

\section{Experiments}
To provide a comprehensive picture of overlapping transformations and DVAE++, we conduct three sets of experiments. In Sec.~\ref{sec:prev_comp} and Sec.~\ref{sec:prev_vae}
we train a VAE with several layers of latent variables with a feed-forward encoder and decoder. This allows to compare
overlapping transformations with previous work on discrete latent variables. In Sec.~\ref{sec:dvae_expr}, we then compare DVAE++ to several baselines.

\begin{table*}
\caption{Overlapping transformations are compared against different single-sample based approaches proposed for training binary latent variable models. 
The performance is measured by 100 importance weighted samples~\cite{burda2015importance}. Mean $\pm$ standard deviation for five runs are reported. Baseline performances are taken from \cite{tucker2017rebar}.} \label{tab:res_sbn}
\small
\centering
\begin{tabular}{ l r r r r r r}
\textbf{MNIST (static)}   & NVIL    & MuProp & REBAR &  Concrete & Joint ELBO & Marg. ELBO \\
\hline
Linear       & -108.35 $\pm$ 0.06 & -108.03 $\pm$ 0.07 & -107.65 $\pm$ 0.08 & \textbf{-107.00 $\pmb{\pm}$ 0.10} & -107.98 $\pm$ 0.10 & -108.57 $\pm$ 0.10 \\
Nonlinear    & -100.00 $\pm$ 0.10 & -100.66 $\pm$ 0.08 & -100.69 $\pm$ 0.08 & -99.54 $\pmb{\pm}$ 0.06 & \textbf{-99.16 $\pmb{\pm}$ 0.12} & \textbf{-99.10 $\pmb{\pm}$ 0.21}  \\
\textbf{OMNIGLOT} & &  &  &  &  & \\
\hline
Linear       & \textbf{-117.59 $\pmb{\pm}$ 0.04} & -117.64 $\pm$ 0.04 & -117.65 $\pm$ 0.04 & -117.65 $\pm$ 0.05 & \textbf{-117.38 $\pmb{\pm}$ 0.08}   & -118.35 $\pm$ 0.06 \\
Nonlinear    & -116.57 $\pm$ 0.08 & -117.51 $\pm$ 0.09 & -118.02 $\pm$ 0.05 & -116.69 $\pm$ 0.08 & \textbf{-113.83 $\pmb{\pm}$ 0.11}  & \textbf{-113.76 $\pmb{\pm}$	0.18} \\
\hline
\end{tabular} 
\end{table*}

\setlength{\tabcolsep}{3pt}   
\begin{table*}
\caption{The performance of the VAE model with an RBM prior trained with the overlapping transformation 
is compared against \cite{rolfe2016discrete} as well as the directed VAE models (Fig.~\ref{fig:sbn}). 
The performance is measured by 4000 importance weighted samples \cite{burda2015importance}. Mean $\pm$ standard deviation for five runs are reported.} \label{tab:res_rbm}
\footnotesize
\centering
\begin{tabular}{ l r r r r | r r r r}
\hline
\multicolumn{1}{c}{} & \multicolumn{4}{c|}{\textbf{MNIST (static)}} & \multicolumn{4}{c}{\textbf{OMNIGLOT}} \\
& RBM (ours) & RBM (Rolfe) & Joint ELBO & Marg. ELBO & RBM (ours) & RBM (Rolfe) & Joint ELBO & Marg. ELBO \\
\hline   
1 Linear    & \textbf{-91.21$\pmb{\pm}$0.11} & -91.55$\pm$0.08 & -106.70$\pm$0.08 &  -106.80$\pm$0.19 & \textbf{-109.66$\pmb{\pm}$0.09} & \textbf{-109.83$\pm$0.17} & -117.62$\pm$0.09 & -117.78$\pm$0.07 \\
2 Linear   & -94.15$\pmb{\pm}$0.45 & \textbf{-91.06$\pmb{\pm}$0.21} & -98.16$\pm$0.11  & -98.56$\pm$0.10 & \textbf{-109.01$\pmb{\pm}$0.45} & -110.35$\pm$0.14 & -111.21$\pm$0.12 & -111.49$\pm$0.08 \\
\hline 
1 Nonlin.    & \textbf{-85.41$\pmb{\pm}$0.04} & -85.57$\pm$0.03 & -95.04$\pm$0.10  & -95.06$\pm$0.08 & \textbf{-102.62$\pmb{\pm}$0.07} & -103.12$\pm$0.06 & -108.77$\pm$0.24 & -108.82$\pm$0.20 \\
2 Nonlin.   & \textbf{-84.27$\pmb{\pm}$0.05} & -84.52$\pm$0.05 & -87.96$\pm$0.13  & -88.23$\pm$0.11 & \textbf{-100.55$\pmb{\pm}$0.05} & -105.60$\pm$0.68 & -103.57$\pm$0.15 & -104.05$\pm$0.22 \\
\hline
\end{tabular}
\end{table*}

\subsection{Comparison with Previous Discrete Latent Variable Models} \label{sec:prev_comp}

We compare overlapping transformations to NVIL~\cite{mnih2014neural}, MuProp~\cite{gu2015muprop}, REBAR~\cite{tucker2017rebar}, and Concrete~\cite{maddison2016concrete} for training discrete single-layer latent variable models. 
We follow the structure used by \cite{tucker2017rebar} in which
the prior distribution and inference model are factorial Bernoulli with 200 stochastic variables. In this setting,
the inference and generative models are either linear or nonlinear functions. In the latter case, two layers 
of deterministic hidden units of the size 200 with tanh activation are used.

We use the settings in \cite{tucker2017rebar} to initialize the parameters, define the model,
and optimize the parameters for the same number of iterations. However, 
\cite{tucker2017rebar} uses the Adam optimizer with $\beta_2=0.99999$ in training. We used Adam with its default parameters except for $\epsilon$ which is set to $10^{-3}$. The learning rate is selected from the set $\{1\cdot 10^{-4}, 5 \cdot 10^{-4}\}$. The inverse temperature $\beta$ for smoothing is annealed linearly during training with initial and final values chosen using cross validation from $\{5, 6, 7, 8\}$ and 
$\{12, 14, 16, 18\}$ respectively. In Table~\ref{tab:res_sbn}, the performance of our model is compared
with several state-of-the-art techniques proposed for training
binary latent models on (statically) binarized MNIST~\cite{salakhutdinov2008dbn} and OMNIGLOT~\cite{lake2015human}. At test time, all models are evaluated in the binary limit ($\beta=\infty$).
Smoothing transformations slightly outperform previous techniques in most cases. In the case of the nonlinear model on OMNIGLOT, the difference is about 2.8 nats. 

\subsection{Comparison with Previous RBM Prior VAE} \label{sec:prev_vae}

Techniques such as KL annealing~\cite{sonderby2016ladder}, batch normalization~\cite{ioffe2015batch}, 
autoregressive inference/prior, and learning-rate decay can significantly improve the performance of a VAE beyond the results reported in Sec.~\ref{sec:prev_comp}.
In this second set of experiments, we evaluate overlapping transformations by comparing the training of a VAE with an RBM prior to the original DVAE \cite{rolfe2016discrete}, both of which include these improvements. 
For a fair comparison, we apply only those techniques that were also used in \cite{rolfe2016discrete}. We examine VAEs with one and two latent layers with feed-forward linear or nonlinear 
inference and generative models. In the one-latent-layer case, the KL term in both our model and \cite{rolfe2016discrete}
reduces to the mean-field approximation.
The only difference in this case lies in the overlapping 
transformations used here and the original smoothing method of \cite{rolfe2016discrete}. 
In the two-latent-layer case, our inference and generative model have the forms depicted in Fig.~\ref{fig:sbn}(d) and Fig.~\ref{fig:sbn}(g).
Again, all models are evaluated in the binary limit at the test time.

Comparisons are reported in Table~\ref{tab:res_rbm}.
For reference, we also provide the performance of the directed VAE models with 
the structures visualized in Fig.~\ref{fig:sbn}(c) to Fig.~\ref{fig:sbn}(f).
Implementation details are provided in Appendix~\ref{app:rbm_details}. 
Two observations can be made from Table~\ref{tab:res_rbm}. First, our smoothing transformation 
outperforms \cite{rolfe2016discrete} in most cases. In some cases the difference is as 
large as 5.1 nats. Second, the RBM prior performs better than a directed prior of the same size. 

\begin{table*}
\caption{DVAE++ compared against different baselines on several datasets. The performance is reported in terms of the log-likelihood values for all the dataset except for CIFAR10, in which \textit{bits per dimension} is reported. In general, DVAE++ with RBM global prior and normal local variables outperforms the baselines.} \label{tab:res_dvae}
\small
\centering
\begin{tabular}{l l l r r r r r}
Latent type & Global & Local  & MNIST (static) & MNIST (dynamic) & OMNIGLOT & Caltech-101 & CIFAR10 \\
\hline
All cont. & Normal & Normal & -79.40 & -78.59 & -92.51 & -82.24 & 3.40 \\
\hline
\multirow{2}{*}{Mixed} & RBM (Rolfe) & Normal    & \textbf{-79.04} & -78.65 & -92.56 & -81.95 & 3.39 \\
& RBM (ours)  & Normal    & -79.17 & \textbf{-78.49} & \textbf{-92.38} & \textbf{-81.88} & \textbf{3.38} \\
\hline
\multirow{2}{*}{All disc.} & RBM (ours)  & Bernoulli & -79.72 & -79.55 & -93.95 & -85.40 & 3.59 \\
& Bernoulli   & Bernoulli & -79.90 & -79.62 & -93.87 & -86.57 & 3.62 \\
\hline
\hline
\multicolumn{3}{l}{Unconditional decoder} & Yes & Yes & No & No & No \\
\end{tabular}
\end{table*}

\begin{figure*}
 \centering
   \subfloat[MNIST]{\includegraphics[height=4.0cm]{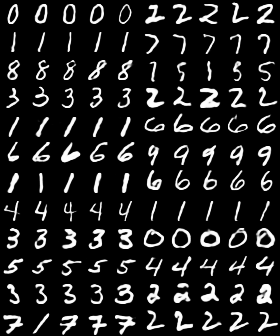}} \hspace{0.1cm}
   \subfloat[OMNIGLOT]{\includegraphics[height=4.0cm]{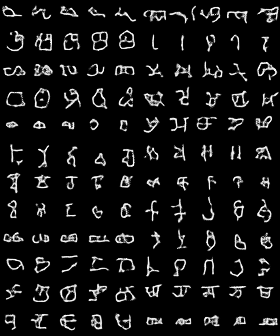}} \hspace{0.1cm}
   \subfloat[Caltech-101]{\includegraphics[height=4.0cm]{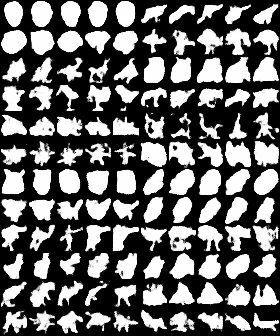}} \hspace{0.1cm}
   \subfloat[CIFAR10]{\includegraphics[height=4.0cm]{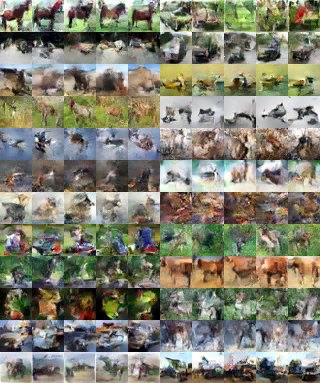}} \hspace{0.1cm}
  \caption{Visualization of samples generated from our model trained on different datasets. In each figure,
  every five successive samples in each row are generated from a fixed sample drawn from the global RBM prior. Our global latent variables typically capture discontinuous global structures such as digit
  classes in MNIST or scene configuration in CIFAR10.}
  \label{fig:vis}
\end{figure*}

\subsection{Experiments on DVAE++} \label{sec:dvae_expr}

Lastly, we explore the performance of DVAE++ for density estimation on 2D images. In addition to statically binarized MNIST and 
OMNIGLOT, we test dynamically binarized MNIST~\cite{lecun1998gradient}
and Caltech-101 silhouettes~\cite{marlin2010inductive}. 
All datasets have $28\times28$ binary pixel images. We use the same architecture for
the MNIST and OMNIGLOT datasets, but because the Caltech-101 silhouettes dataset 
is smaller, our model easily overfits. Consequently, we use a shallower architecture for Caltech-101.
We also evaluate DVAE++ on the CIFAR10 dataset, which consists of $32\times32$ pixel natural images. 
Appendix~\ref{app:impDetails} lists the details of our architecture for different datasets.

Our goal is to determine whether we can use overlapping transformations to train a convolutional VAE with an RBM prior, and whether
the RBM prior in DVAE++ captures global discrete hidden factors.
In addition to DVAE++ (which uses binary global latent variables and continuous local latent variables), four different 
baselines are introduced by modifying the global and local distributions.
These baselines are listed in Table~\ref{tab:res_dvae}. For RBM~(Rolfe), the spike-and-exp smoothing transformation
is used and the ELBO is optimized using the derivation supplied in \cite{rolfe2016discrete}. For Bernoulli latent variables, we used
the marginal distributions proposed in Sec.~\ref{sec:cont_relax}.
For all the models, we used 16 layers of local latent variables each with 32 random variables at each spatial location.
For the RBM global variables, we used 16 binary variables for all the binary datasets and 
128 binary variables for CIFAR10. We cross-validated the number of the hierarchical layers 
in the inference model for the global variables from the set $\{1, 2, 4\}$. We used an unconditional decoder 
(i.e., factorial $p(\x|\bzeta, \h)$) for the MNIST datasets. 
We measure performance by estimating test set log-likelihood (again, according to the binary model) with 4000 importance weighted samples.
Appendix~\ref{app:ablation} presents additional ablation experiments. 

Table~\ref{tab:res_dvae} groups the baselines into three categories: all continuous latent, discrete global and continuous local (mixed), and all discrete. Within the mixed 
group, DVAE++ with RBM prior generally outperforms the same model trained with \cite{rolfe2016discrete}'s. Replacing the continuous normal local variables with Bernoulli variables does not dramatically hurt the performance. For example, in the case of statically and dynamically binarized MNIST dataset, we achieve $-79.72$ and $-79.55$ respectively with unconditional 
decoder and $3.59$ on CIFAR10 with conditional decoder. To the best of our knowledge these are the best reported results on these datasets with binary latent variables. 
Samples generated from DVAE++ are visualized in Fig.~\ref{fig:vis}. As shown, the discrete global prior clearly captures discontinuous latent factors such as digit category or scene configuration.

DVAE++ results are comparable to current state-of-the-art convolutional latent variable models such as VampPrior~\cite{tomczak2017vae} and variational lossy autoencoder (VLAE)~\cite{chen2016variational}. We note two features of these models that may offer room for further improvement for DVAE++. First, the conditional decoder used here makes independence 
assumptions in each scale, whereas the state-of-the-art techniques are based on PixelCNN~\cite{van2016pixel}, which assumes full autoregressive dependencies. Second, methods such as VLAE use normalizing flows for flexible inference models that reduce the KL cost on the convolutional latent variables. Here, the independence assumption in each local group in
DVAE++ can cause a significant KL penalty. 

\section{Conclusions}
We have introduced a new family of smoothing transformations consisting 
of a mixture of two overlapping distributions and have demonstrated that
these transformations can be used for training latent variable models
with either directed or undirected priors. Using variational bounds
derived for both cases, we developed DVAE++ having a global RBM prior
and local convolutional latent variables. All experiments used exponential mixture components, but it
would be interesting to explore the efficacy of other choices.

\bibliography{generative.bib}
\bibliographystyle{icml2018}

\clearpage
\appendix
\section{Overlapping Transformation with the Mixture of Exponential Distributions} \label{app:overlap}

The CDF for each conditional distribution is given by
\begin{align*}
F_{r(\zeta |z =0)}(\zeta) &=  \frac{1 - e^{-\beta \zeta}}{1 - e^{-\beta}} \\
F_{r(\zeta |z =1)}(\zeta) &=  \frac{e^{\beta (\zeta - 1)} - e^{-\beta}}{1 - e^{-\beta}}.
\end{align*}
To simplify notation, the mean of the Bernoulli distribution $q(z=1|x)$ is denoted by $q$. The CDF for the 
mixture $q(\zeta|x) = \sum_z r(\zeta|z) q(z|x)$ is
\begin{eqnarray*}
F_{q(\zeta|x)}(\zeta) &=& \frac{1 - q }{1 - e^{-\beta}} \left[1 - e^{-\beta \zeta}\right] +  \\
&&\frac{q}{1 - e^{-\beta}} \left[ e^{\beta (\zeta - 1)} - e^{-\beta} \right].
\end{eqnarray*}
Defining $m \equiv e^{-\beta \zeta }$ and $d \equiv e^{- \beta}$, the inverse CDF
is found by solving $F_{q(\zeta)}(\zeta) -\rho = 0$ which gives rise to the quadratic equation
\begin{eqnarray*}
m^2 + \underbrace{ \left[-1 + \frac{\rho + d(q - \rho)}{1 - q} \right]}_{b} m \underbrace{- \frac{qd}{1-q}}_c  &=& 0,
\end{eqnarray*}
which has solutions $m = (-b \pm \sqrt{b^2 - 4c})/2$. Since $-4c\geq 0$ 
(as $q \geq 0$ and $d > 0$), there are two real solutions. Further, $m = (-b + \sqrt{b^2 - 4c})/2$ is the valid solution since 
$m$ must be positive (recall $m=e^{-\beta\zeta}$) and $\sqrt{b^2 - 4c} \geq |b|$.
Lastly, the inverse CDF is obtained using $\zeta = -\log{m}/\beta$. The inverse CDF is a differentiable mapping from uniform samples 
$\rho \sim U(0,1)$ to samples from $q(\zeta |x)$.

\section{Overlapping Transformation with the Mixture of Logistic Distributions} \label{app:otherOverlap}

The Dirac $\delta$ distribution can be approximated by a normal distribution whose variance approaches to zero. We 
use this observation to define a smoothing transformation where each $r(\zeta | z)$ is modeled with a Normal distribution 
(with $\zeta\in\mathbb{R})$:
\begin{eqnarray*}
r(\zeta | z = 0) &=& \mathcal{N}(\zeta | 0, \sigma^2)\\
r(\zeta | z = 1) &=& \mathcal{N}(\zeta | 1, \sigma^2).
\end{eqnarray*}
The resulting mixture $q(\zeta|x) = \sum_z r(\zeta|z) q(z|x)$ converges to a Bernoulli distribution as $\sigma$ goes to 0, but 
its CDF (which is a mixture of error functions) cannot be inverted in closed form.

To derive a Normal-like distribution with an invertible CDF and support $\zeta\in \mathbb{R}$, 
we define a smoothing transformation using the logistic distribution
\begin{eqnarray*}
r(\zeta | z = 0) &=& \mathcal{L}(\zeta | \mu_0, s)\\
r(\zeta | z = 1) &=& \mathcal{L}(\zeta | \mu_1, s) 
\end{eqnarray*}
where
\begin{equation*}
 \mathcal{L}(\zeta | \mu, s) = \frac{e^{- \frac{\zeta - \mu}{s}}}{s (1 + e^{- \frac{\zeta - \mu}{s}})^2}.
\end{equation*}
For the mixture distribution\footnote{$q$ is again shorthand for $q(z=1|x)$.}
\begin{equation*}
 q(\zeta | x) = (1 - q) \mathcal{L}(\zeta, \mu_0, s) + q \mathcal{L}(\zeta, \mu_1, s),
\end{equation*}
the inverse CDF is derived by solving
\begin{eqnarray*}
 F_{q(\zeta)}(\zeta) = \frac{1-q}{1 + e^{ - \frac{\zeta - \mu_0}{s}} } + \frac{q}{1 + e^{ - \frac{\zeta - \mu_1}{s} }} = \rho \\
 \frac{  (1 - q)(1 + e^{ - \frac{\zeta - \mu_1}{s}})  + q(1 + e^{ - \frac{\zeta - \mu_0}{s}}) }{ (1 + e^{ - \frac{\zeta - \mu_1}{s}})  (1 + e^{ - \frac{\zeta - \mu_0}{s}}) } = \rho.
\end{eqnarray*}
Defining $m \equiv e^{-\zeta/s}$, $d_0 \equiv e^{\mu_0/s}$, and $d_1=e^{\mu_1/s}$ yields a quadratic in $m$
\begin{eqnarray*}
\underbrace{\rho d_0 d_1}_a m^2 + \underbrace{[\rho(d_0 + d_1) - d_0 q - d_1 (1-q)]}_bm + \underbrace{\rho - 1}_c = 0
\end{eqnarray*}
which has the valid solution
\begin{equation*}
 m^* = \frac{-b + \sqrt{b^2 - 4ac}}{2a}.
\end{equation*}
This gives the inverse CDF as $F^{-1}_{q(\zeta|x)}(\rho) = -s \log{m^*}$. When $s$ is very small and $\mu_1 > \mu_0$, 
$d_1$ is suceptible to overflow. A numerically stable solution can be obtained by applying the change of variable $m'=\sqrt{d_0d_1}m$.

\section{Visualization of Inverse CDFs} \label{app:invCDFViz}
\begin{figure*}
  \centering
    \subfloat{\includegraphics[scale=0.35, trim={0.4cm 0 0.4cm 0},clip]{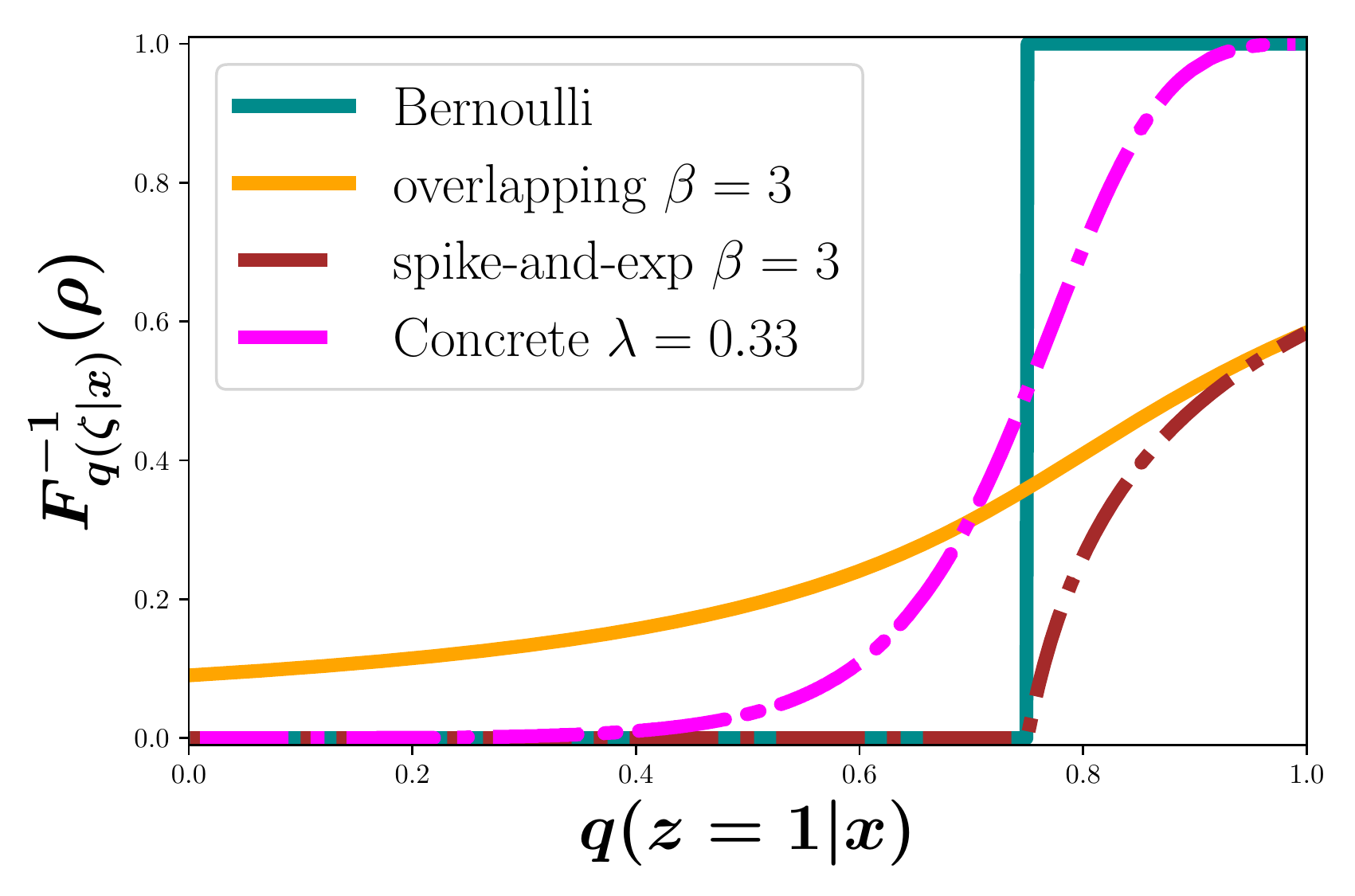}}
    \subfloat{\includegraphics[scale=0.35, trim={0.4cm 0 0.4cm 0},clip]{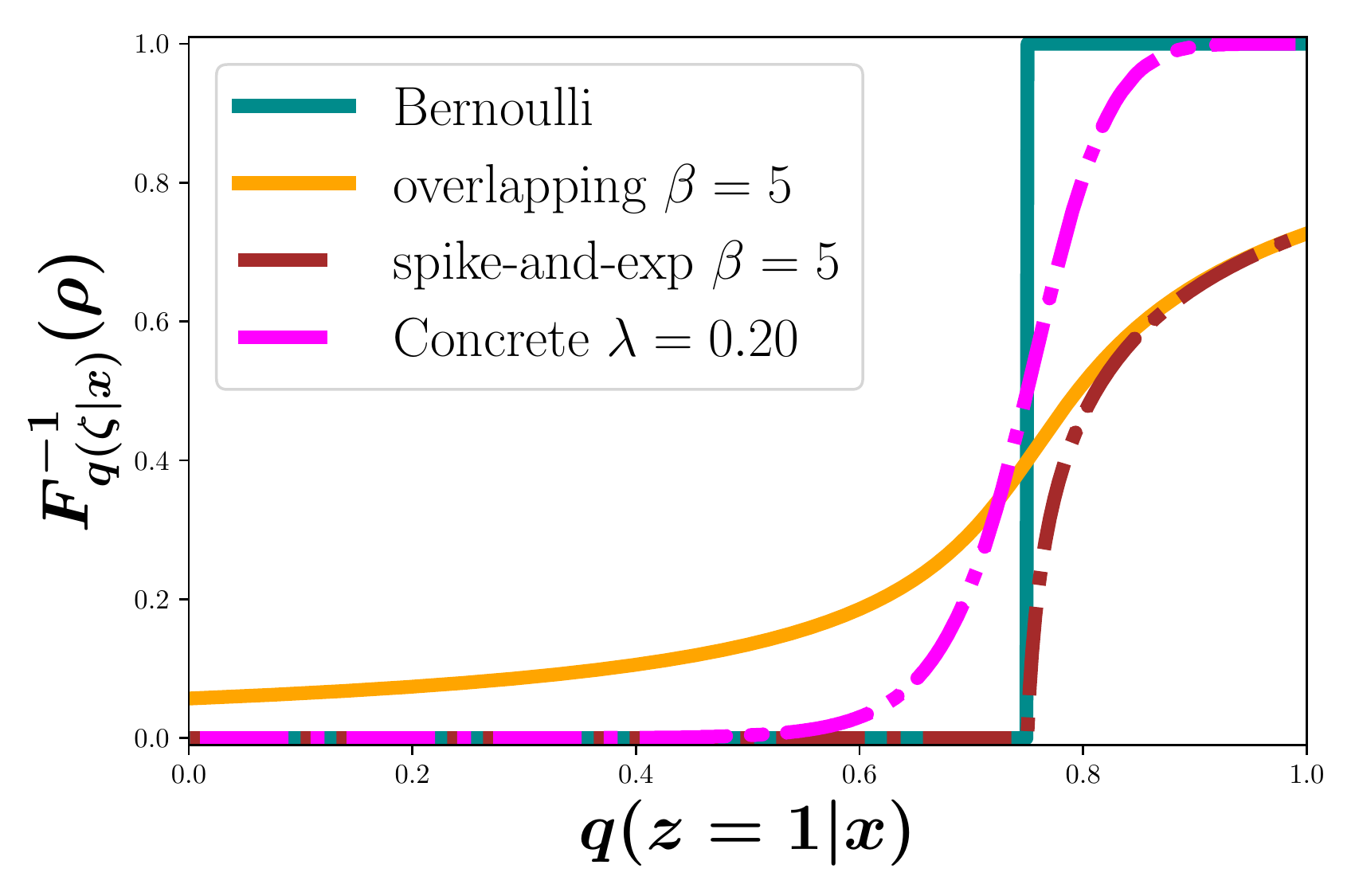}}
    \subfloat{\includegraphics[scale=0.35, trim={0.4cm 0 0.4cm 0},clip]{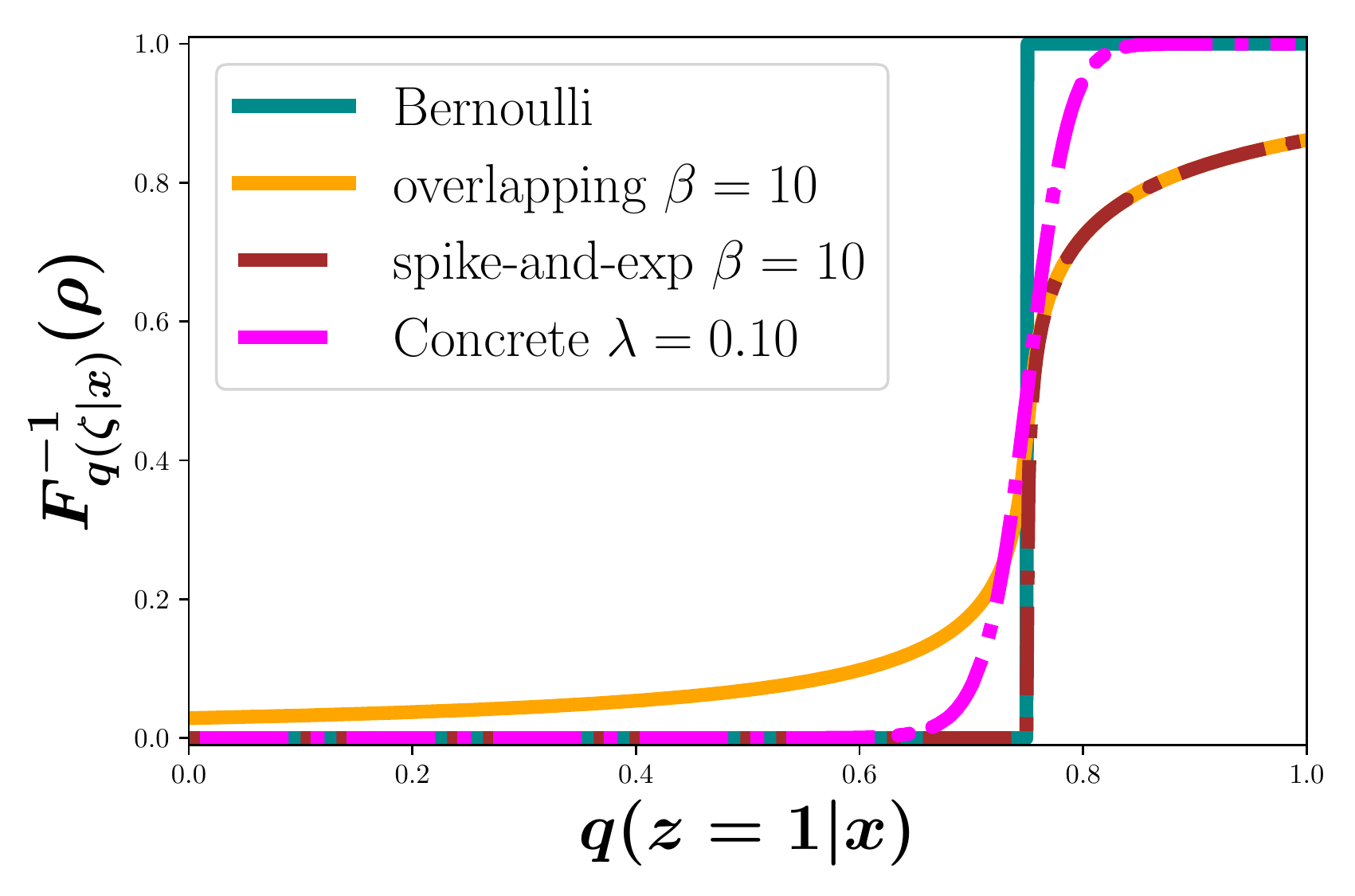}}
    \caption{Visualization of inverse CDF as a function of $q(z=1|x)$ at $\rho=0.25$ for different smoothing transformations with
    three different temperatures ($\lambda$) and inverse temperatures ($\beta$). We have selected temperature values that are often 
    used in practice.
    }
     \label{fig:comp_inv_cdf}
\end{figure*}
To provide insight into the differences between overlapping transformations, Concrete~\cite{maddison2016concrete, jang2016categorical}, and 
spike-and-exponential~\cite{rolfe2016discrete} smoothing, Fig.\ref{fig:comp_inv_cdf} visualizes the inverse CDFs at different temperatures. 
In cases where Concrete and spike-and-exponential have small gradients with respect to $q(z=1|x)$ (thereby slowing learning), overlapping transformations 
provide a larger gradient signal (for faster learning).

\section{Joint versus Marginal ELBOs} \label{app:elbos}

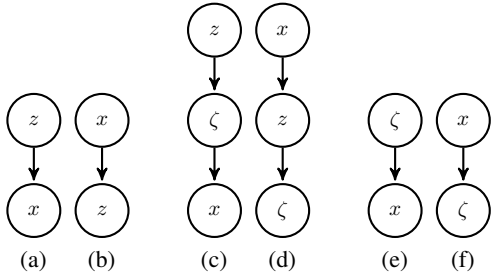
\begin{figure}
 \centering
 \subfloat[]{
  \begin{tikzpicture}[->,>=stealth',shorten >=1pt,auto,node distance=1.5cm, thick, scale=0.80]
      \tikzstyle{every state}=[fill=white,draw=black,text=black, transform shape]
      \node[state] 			 (z1)                   {$z$};
      \node[state] 			 (x)    [below of=z1]   {$x$};
      \path (z1)        edge  (x);
  \end{tikzpicture}  
  }
   \subfloat[]{
 \begin{tikzpicture}[->,>=stealth',shorten >=1pt,auto,node distance=1.5cm, thick, scale=0.80]
      \tikzstyle{every state}=[fill=white,draw=black,text=black, transform shape]
      \node[state] 			   (x)                    {$x$};
      \node[state] 			   (z1)   [below of=x]    {$z$};
      \path (x)         edge               (z1);
  \end{tikzpicture}  
  } \hspace{0.5cm}
  \subfloat[]{
  \begin{tikzpicture}[->,>=stealth',shorten >=1pt,auto,node distance=1.5cm, thick, scale=0.80]
      \tikzstyle{every state}=[fill=white,draw=black,text=black, transform shape]
      \node[state] 		(z1)                   {$z$};
      \node[state] 		(zeta1)   [below of=z1]   {$\zeta$};
      \node[state] 			 (x)    [below of=zeta1]   {$x$};
      \path (z1)        edge             (zeta1);
      \path (zeta1)        edge             (x);
  \end{tikzpicture}  
  } 
  \subfloat[]{
 \begin{tikzpicture}[->,>=stealth',shorten >=1pt,auto,node distance=1.5cm, thick, scale=0.80]
      \tikzstyle{every state}=[fill=white,draw=black,text=black, transform shape]
      \node[state] 	(x)                       {$x$};
      \node[state] 	(z1)     [below of=x]    {$z$};
      \node[state] 	(zeta1)  [below of=z1]   {$\zeta$};
      \path (x)         edge               (z1);
      \path (z1)        edge               (zeta1);
  \end{tikzpicture}  
  } \hspace{0.5cm}
  \subfloat[]{
  \begin{tikzpicture}[->,>=stealth',shorten >=1pt,auto,node distance=1.5cm, thick, scale=0.80]
      \tikzstyle{every state}=[fill=white,draw=black,text=black, transform shape]
      \node[state] 			 (z1)                   {$\zeta$};
      \node[state] 			 (x)    [below of=z1]   {$x$};
      \path (z1)        edge  (x);
  \end{tikzpicture}  
  }
   \subfloat[]{
 \begin{tikzpicture}[->,>=stealth',shorten >=1pt,auto,node distance=1.5cm, thick, scale=0.80]
      \tikzstyle{every state}=[fill=white,draw=black,text=black, transform shape]
      \node[state] 			   (x)                    {$x$};
      \node[state] 			   (z1)   [below of=x]    {$\zeta$};
      \path (x)         edge               (z1);
  \end{tikzpicture}  
  }
  \caption{(a) A generative model with binary latent variable $z$. (b) The corresponding inference model. In (c) and (d), the continuous $\zeta$ is introduced and dependency on $z$ is transferred to dependency on $\zeta$. In (e) and (f)
  the binary latent variable $z$ is marginalized out.}
  \label{fig:directed}
\end{figure}
We have presented two alternative ELBO bounds, one based on a joint inference model $q(\bzeta_1,\bzeta_2,\z_1,\z_2|\x)$ and the other based on $q(\bzeta_1,\bzeta_2|\x)$ obtained by marginalizing the discrete variables. Here, we show that variational bound obtained with the marginal model is tighter.

For simplicity we consider a model with only one latent variable. Figs.~\ref{fig:directed}(a) and (b) visualize the generative and inference models. Figs.~\ref{fig:directed}(c) and (d) show the joint models, and Figs.~\ref{fig:directed}(e) and (f) show the marginal models.
The respective variational bounds are
\begin{align*}
L_1(\x)  &= \log p(x) - \E_{q(\z, \bzeta|\x)} \left[ \frac{q(\z, \bzeta|\x)}{p(\z, \bzeta|\x)}  \right]
\intertext{and}
L_2(\x) & = \log p(\x) - \E_{q(\bzeta|\x)} \left[ \log \frac{q(\bzeta|\x)}{p(\bzeta|\x)} \right].
\end{align*}

Subtracting we find 
\begin{align*}
L_2(\x) - L_1(\x) = & \ \E_{q(\bzeta,\z|\x)} \left[ \log \frac{q(\bzeta,\z|\x)}{p(\bzeta,\z|\x)} \right] - \\
 & \ \E_{q(\bzeta|\x)}\left[ \log \frac{q(\bzeta|\x)}{p(\bzeta|\x)} \right] \\
 = & \ \E_{q(\bzeta,\z|\x)} \biggl[ \log \frac{q(\bzeta,\z|\x)}{p(\bzeta,\z|\x)} -
 \log \frac{q(\bzeta|\x)}{p(\bzeta|\x)} \biggr] \\
 = & \ \E_{q(\bzeta,\z|\x)} \biggl[ \log \frac{q(\z|\bzeta,\x)}{p(\z|\bzeta,\x)} \biggr] \\
 =& \ \E_{q(\bzeta|\x)} \bigl[\KL\bigl( q(\z|\bzeta,\x) \| p(\z|\bzeta,\x) \bigr) \bigr],
\end{align*}
which is clearly positive since $\KL(\cdot\|\cdot)\ge 0$. Thus, $L_2(\x)$, the marginal ELBO, provides a tighter bound.

\section{Adding the Gradient of $\log Z$ to the Objective Function} \label{app:gradZ}

For training the DVAE++ model with an RBM prior, the gradient of $\log Z$ is needed for each parameter update.
Since $\log Z$ only depends on the prior parameters $\theta=\{\a_1, \a_2, \W\}$, its gradient is
\begin{flalign*}
\frac{\partial \log Z}{\partial \theta} &= \frac{\partial}{\partial \theta} \log \sum_{\z_1, \z_2} e^{-E_{\theta}(\z_1, \z_2)}   \\
&= - \frac{\sum_{\z_1, \z_2} e^{-E_{\theta}(\z_1, \z_2)} \frac{\partial E_{\theta}(\z_1, \z_2)}{\partial \theta}}{\sum_{\z'_1, \z'_2} e^{-E_{\theta}(\z'_1, \z'_2)}} \\
&= -\sum_{\z_1, \z_2} p_\theta(\z_1, \z_2) \frac{\partial E_{\theta}(\z_1, \z_2)}{\partial \theta} \\
&= - \E_{p_\theta(\z_1, \z_2)}\left[ \frac{\partial E_{\theta}(\z_1, \z_2)}{\partial \theta}  \right].
\end{flalign*}
This expectation is estimated using Monte Carlo samples from the RBM. We maintain persistence chains and run block Gibbs updates for 
a fixed number of iterations ($40$) to update the samples after each parameter update. This approach is known as 
persistent contrastive divergence (PCD) \cite{younes1989parametric, tieleman2008training}. 

Instead of manually coding the gradient of the negative energy function for each
sample and modifying the gradient of whole objective function, we compute the negative average energy on $L$ samples (indexed by $l$) generated from PCD chains 
($-\sum_{l=1}^L E_{\theta}(\z^{(l)}_1, \z^{(l)}_2)/L$ where $\z^{(l)}_1, \z^{(l)}_2 \sim p_\theta(\z_1, \z_2)$). 
This gives a scalar tensor whose gradient is the sample-based approximation to $\partial Z/\partial \theta$. By adding this tensor to the objective 
function, an automatic differentiation (AD) library backpropagates through this tensor and computes the appropriate gradient 
estimate. Note that an AD library cannot backpropagate the gradients through the samples generated from the PCD computation graph because of the discrete 
nature of the process. However, one can use stop gradient commands on the samples to prevent unnecessary AD operations.

\section{Implementation Details for the RBM Prior Experiments} \label{app:rbm_details}

In this section, we summarize the implementation details for the experiments reported in Sec.\ref{sec:dvae_expr} on the RBM prior VAE. 
During training, the KL term is annealed linearly from 0 to 1 in 300K iterations. The learning rate starts at $3\cdot 10^{-3}$ and is multiplied by 
0.3 at iterations 600K, 750K, and 950K. 
Batch normalization is used for the nonlinear deterministic hidden layers. Training runs for 1M iterations 
with batch size of 100 and the performance is measured using 4000-sample importance weight estimation of log-likelihood. 
$\beta$ is fixed during training for all methods and is cross-validated from the set $\{5,6,8,10\}$. 
Nonlinear models use two hidden layers of size 200 with tanh activations. 
There are 200 stochastic latent variables in each layer. 

\section{Implementation Details for the DVAE++ Experiments} \label{app:impDetails}
\begin{figure*}
 \centering
   \subfloat[encoder]{
  \begin{tikzpicture}[->,>=stealth',shorten >=1pt,auto,node distance=1.0cm, thick, scale=0.85]
       \tikzstyle{every state}=[fill=white,draw=black,text=black, transform shape, shape=rectangle, minimum size=0.7cm, ultra thick]
       \tikzstyle{line} = [draw, -latex']
       \node at (0,0) [state, shape=circle]                                      (x)          {$\x$};
       \node [state, draw=teal, right=0.4 of x] 		                           (c1)        {down 1};
       \node [state, draw=teal, below= 0.4 cm of c1] 	               	 (c2)        {down 2};
       \coordinate[right= 0.5cm of c2]                                           (c2a);
       \node [state, draw=orange, right=0.7cm of c2a] 	 (fc1)        {$q(\z_1|\x)$};
       \node [state, draw, right=0.7cm of fc1, shape=circle] 		     (z1)        {$\bzeta_1$};
       \node [draw, below= 0.5cm of c2a, shape=circle, scale=0.5]      (p1)        {$+$};
        \node [state, draw=orange, below=0.8 cm of fc1] 		            (fc2)        {$q(\z_2|\x, \bzeta_1)$};
        \node [state, draw,  right=0.5cm of fc2, shape=circle] 		     (z2)        {$\bzeta_2$};
        \node [draw, right= 0.5cm of z1, shape=circle, scale=0.5]   (p2)        {$+$};
        \node [state, draw=teal, right=0.5cm of p2] 		                        (u1)        {upsample};
        \node [draw, right=0.5cm of u1,  shape=circle, scale=0.5] 	     (p3)        {$+$};
        \coordinate[right= 0.3cm of p3]                                              (p3a);
        \node [state, draw=teal, right=0.7 of p3a] 		                           (qh1)        {$q(\h_1|\x, \bzeta)$};
        \node [state, draw, right= 0.7cm of qh1, shape=circle] 		     (h1)        {$\h_1$};
        \node [draw, below=0.5 of p3a, shape=circle, scale=0.5] 		 (p4)        {$+$};
        \node [state, draw=teal, below=0.8cm of qh1] 		                           (qh2)        {$q(\h_2|\x, \bzeta, \h_1)$};
        \node [state, draw, right= 0.4cm of qh2, shape=circle] 		     (h2)        {$\h_2$};
       \path[->] (x) edge (c1);
       \path [line] (c1) edge (c2);
       \path [line] (c2) -- (fc1);
       \path [line] (fc1) edge (z1);
       \path [line] (z1) |-  (p1);
       \path [line] (c2) -| (p1);
       \path [line] (p1) |-  (fc2);
       \path [line] (fc2) edge (z2);
       \path [line] (z1) edge  (p2);
       \path [line] (z2) -| (p2);
       \path [line] (p2) edge (u1);
       \path [line] (u1) edge (p3);
       \path [line] (c1) -| (p3);
       \path [line] (p3) edge (qh1);
       \path [line] (qh1) edge (h1);
       \path [line] (p3) -| (p4);
       \path [line] (h1) |- (p4);
       \path [line] (p4) |- (qh2);
       \path [line] (qh2) edge (h2);
  \end{tikzpicture}  
  } \\
     \subfloat[prior]{
  \begin{tikzpicture}[->,>=stealth',shorten >=1pt,auto,node distance=1.0cm, thick, scale=0.85]
       \tikzstyle{every state}=[fill=white,draw=black,text=black, transform shape, shape=rectangle, minimum size=0.7cm, ultra thick]
       \tikzstyle{line} = [draw, -latex']
       \node [state, shape=circle]                                      (z1)          {$\z_1$};
       \node [state, below of=z1, shape=circle]                  (z2)          {$\z_2$};
       \node [state, shape=circle, right=0.5 cm of z1]               (zeta1)          {$\zeta_1$};
       \node [state, shape=circle, right=0.5 cm of z2]                  (zeta2)       {$\zeta_2$};
       \draw [dashed, rounded corners=.3cm] (-0.5,0.5) rectangle (0.5,-1.5);
       \coordinate    (z1z2a) at (1.7, 0.0);
       \node [draw, right=0.4 of zeta1, shape=circle, scale=0.5] 		    (p0)        {$+$};
       \node [state, draw=teal, right=0.4 of p0] 		       (u1)        {upsample};
       \coordinate  [right=0.7cm of u1] (u1a);
       \coordinate  [above=0.6cm of u1a] (u1b);
       \node [state, draw=teal, right=1.0 of u1a] 		       (ph1)        {$p(\h_1|\bzeta)$};
       \node [state, draw, right= 0.6cm of ph1, shape=circle] 		     (h1)        {$\h_1$};
       \node [draw, below=0.4 of u1a, shape=circle, scale=0.5] 		    (p1)        {$+$};
       \node [state, draw=teal, below=0.6cm of ph1] 		       (ph2)        {$p(\h_2|\bzeta, \h_1)$};
       \node [state, draw, below=0.6cm of h1, shape=circle] 		     (h2)        {$\h_2$};
       \node [draw, right=0.4 of h1, shape=circle, scale=0.5] 		    (p2)        {$+$};
       \node [right=0.4 of p2] 		                                                 (dd)        {to decoder};
       \path [line] (z1) edge (zeta1);
       \path [line] (z2) edge (zeta2);
       \path[-] (z1) edge (z2);
       \path [line] (zeta1) edge (p0);
       \path [line] (zeta2) -| (p0);
       \path [line] (p0) -- (u1);
       \path [line] (u1) -- (ph1);
       \path [line] (ph1) edge (h1);
       \path [line] (h1) |- (p1);
       \path [line] (u1) -| (p1);
       \path [line] (p1) |- (ph2);
       \path [line] (ph2) -- (h2);
       \path [line] (h2) -| (p2);
       \path [line] (u1) -| (u1b) -| (p2);
       \path [line] (h1) -- (p2);
       \path [line] (p2) edge (dd);
  \end{tikzpicture}
  \hspace{1cm}
  \begin{tikzpicture}[->,>=stealth',shorten >=1pt,auto,node distance=1.0cm, thick, scale=0.7]
       \tikzstyle{every state}=[fill=white,draw=black,text=black, transform shape, shape=rectangle, minimum size=0.4cm, ultra thick]
       \tikzstyle{line} = [draw, -latex']
       \node [draw, shape=circle, scale=0.5] 		           (p)         {$+$};
       \node [right=0.4cm of p]                                    (pp)        {\footnotesize concatenate};
       \node [state,below=0.1 cm of p, shape=circle, scale=0.9]                       (z)          {$\z$};
       \node [right=0.4cm of z]                                    (zz)        {\footnotesize random variable};
       \node [state, draw=teal, below=0.1 cm of z] 		   (r)        {$\quad \quad$};
       \node [right=0.25cm of r]                                    (rr)        {\footnotesize residual network};
       \node [state, draw=orange, below=0.1 cm of r]               (f)        {$\quad \quad$};
       \node [right=0.25cm of f]                                    (ff)        {\footnotesize fully connected network};
       \node [state, draw=magenta, below=0.1 cm of f] 		   (c)        {$\quad \quad$};
       \node [right=0.25cm of c]                                    (cc)        {\footnotesize 2D convolution};
       
  \end{tikzpicture}  
  }
  \\
       \subfloat[decoder]{
  \begin{tikzpicture}[->,>=stealth',shorten >=1pt,auto,node distance=1.0cm, thick, scale=0.85]
       \tikzstyle{every state}=[fill=white,draw=black,text=black, transform shape, shape=rectangle, minimum size=0.7cm, ultra thick]
       \tikzstyle{line} = [draw, -latex']
	\node                             (dd)        {from prior};
	\node [state, draw=teal, right=0.4 of dd] 		                           (c3)        {context};
	\node [state, draw=magenta, above=0.4 of c3] 		                           (px1)        {$p(\x_{4\times4}|\bzeta, \h)$};
	\node [draw, right=0.6 of c3, shape=circle, scale=0.5] 		                   (p1)        {$+$};
	\node [state, draw=teal, below=0.4 of p1, scale=0.9] 		                   (input1)        {input $4\times4$};
	\node [state, below=0.4 of input1, shape=circle, scale=0.8]                        (x1)          {$\x_{4\times4}$};
	\node [state, draw=teal, right=0.5 of p1] 		                           (deconv1)        {upsample 1};
	\node [state, draw=magenta, above=0.4 of deconv1] 		                   (px2)        {$p(\x_{8\times8}|\bzeta, \h)$};
	\node [draw, right=0.6 of deconv1, shape=circle, scale=0.5] 		           (p2)        {$+$};
	\node [state, draw=teal, below=0.4 of p2, scale=0.9] 		                   (input2)     {input $8\times8$};
	\node [state, below=0.4 of input2, shape=circle, scale=0.8]                        (x2)          {$\x_{8\times8}$};
	\node [state, draw=teal, right=0.5 of p2] 		                           (deconv2)        {upsample 2};
	\node [state, draw=magenta, above=0.4 of deconv2] 		                   (px3)        {$p(\x_{16\times16}|\bzeta, \h)$};
	\node [draw, right=0.6 of deconv2, shape=circle, scale=0.5] 		           (p3)        {$+$};
	\node [state, draw=teal, below=0.4 of p3, scale=0.9] 		                   (input3)     {input $16\times16$};
	\node [state, below=0.4 of input3, shape=circle, scale=0.65]                       (x3)          {$\x_{16\times16}$};
	\node [state, draw=teal, right=0.5 of p3] 		                           (deconv3)        {upsample 3};
	\node [state, draw=magenta, above=0.4 of deconv3] 		                   (px4)        {$p(\x_{32\times32}|\bzeta, \h)$};
	\path [line] (dd) edge (c3);
	\path [line] (c3) edge (px1);
	\path [line] (c3) -- (p1);
	\path [line] (c3) -- (p1);
	\path [line] (input1) -- (p1);
	\path [line] (x1) edge (input1);
	\path [line] (p1) edge (deconv1);
	\path [line] (deconv1) edge (px2);
	\path [line] (deconv1) -- (p2);
	\path [line] (input2) -- (p2);
	\path [line] (x2) edge (input2);
	\path [line] (p2) edge (deconv2);
	\path [line] (deconv2) edge (px3);
	\path [line] (deconv2) edge (p3);
	\path [line] (input3) -- (p3);
	\path [line] (x3) edge (input3);
	\path [line] (p3) edge (deconv3);
	\path [line] (deconv3) edge (px4);
  \end{tikzpicture}  
  }
  \caption{The DVAE++ architecture is divided into three parts: a) the inference network or encoder that represents $q(\z, \bzeta, \h | \x)$, b) the prior network that models $p(\z, \bzeta, \h)$, and c) the decoder network that implements $p(\x| \bzeta, \h)$. Each part consists of different modules colored differently based on their type. The specific detail for each module is listed in Table~\ref{tab:net_spec}.}
  \label{fig:net}
\captionof{table}{The architecture specifics for each module in DVAE++ for different datasets. The numbers correspond to the number of filters used in residual and convolutional blocks or the number of units used in the fully connected layers starting from the first hidden layer up to the last layer in each module. The arrows \dd $\ $ and \uu $\ $ in front of each number indicate that the corresponding block is downsampling or upsampling its input. For Binarized MNIST and MNIST,
we used an identical architecture to OMNIGLOT except that the autoregressive connections are disabled when forming $p(\x | \bzeta, \h)$ in the decoder.} \label{tab:net_spec}
\centering
\begin{tabular}{ l|c|c|c|c}
Module & Type & OMNIGLOT & Caltech-101 & CIFAR10 \\
\hline
down 1 									 & residual& 32\dd, 32, 64\dd, 64, 128\dd    &  8\dd, 8\dd               &  64\dd, 64, 256\dd, 256 \\
down 2 									 & residual & 256\dd, 512\dd                        &  8\dd, 16\dd, 16\dd  & 512\dd, 512, 1024\dd, 1024\dd \\
$q(\z_i|\x, \bzeta_{<i})$  			 	 & fully connected&      4                               &          4                      &      16                           \\
upsample 									 & residual &  128\uu, 128\uu                       &32\uu, 32\uu, 32\uu   &  128\uu, 64\uu, 32\uu  \\
$q(\h_i|\x, \bzeta, \h_{<i})$   & residual & 32, 32, 32, 32, 64                    &  32, 32, 32, 32, 64     &      32, 64        \\
$p(\h_i|\bzeta, \h_{<i})$ 		 & residual &  32, 32, 64                               &   32, 32, 32, 32, 64    &       32, 32, 32, 32, 64       \\
context						 			 & residual &      16         							    &               16                &     256      \\
upsample 1						 	     & residual &     16\uu, 16,                           &          8, 8                   &       128, 128,      \\
upsample 2						 	     & residual &     16\uu, 8                              &          8\uu, 4              &         64\uu, 64    \\
upsample 3						 	     & residual &      8\uu                                   &           4\uu                 &       32\uu            \\
$p(\x_i | \bzeta, \h, \x_{<i}) $ & convolutional &              1                       &                 1                &     100    \\
input $h\times w$ & residual    &              32                       &                 32                &     32    \\

\end{tabular}
\end{figure*}

The network architecture visualized in Fig.~\ref{fig:net} is divided into three parts: i) the inference network or encoder that represents $q(\z, \bzeta, \h | \x)$, ii)
the prior network that models $p(\z, \bzeta, \h)$, and iii)
the decoder network that implements $p(\x| \bzeta, \h)$. Our architecture consists of many modules that may differ for different datasets. The details here describe the 
network architecture used for CIFAR10 data. Table~\ref{tab:net_spec} lists the specifics of the networks for all the datasets.

For the encoder side, ``down 1'' denotes a series of downsampling residual blocks that extract convolutional features from the input image. The output of this network is a feature of size $8\times8\times256$ (expressed by height $\times$ width $\times$ depth). The module ``down 2'' denotes another residual network that takes the output of ``down 1'' and progressively reduces the spatial dimensions of the feature maps until it reaches to a feature map of size $1\times1\times1024$. This feature map is flattened and is iteratively fed to a series of fully connected networks that model $q(\z_i | \x, \bzeta_{<i})$. In addition to the feature map, each network accepts the concatenation of samples drawn from the smoothed variables. These networks have identical architecture with no parameter sharing and all model factorial Bernoulli distributions in their output.

The concatenation of samples from smoothed variables $\bzeta_i$ is fed to ``upsample'', a residual network that upsamples its 
input to $8\times8\times\times32$ dimensions using transposed convolutions. These features are concatenated with the feature
map generated by ``down 1''.  The concatenated feature is then iteratively fed to a series of residual network 
that defines $q(\h_j | \x, \bzeta, \h_{<j})$. Each network accepts the concatenation samples from local variables 
in the previous group ($\h_{<j}$) in addition to the concatenated feature. When the local latent variables are modeled 
by normal distributions, these networks return mean and logarithm of the standard deviation of the elements in $\h_j$ similar to the original VAE \cite{kingma2014vae}.

In the prior network, the same ``upsample'' network defined above is used to scale-up the global latent variables to the intermediate scale ($8\times8\times32$). 
Then, the output of this network is fed to a set of residual networks that defines $p(\h_j | \bzeta, \h_{<j})$ one at a time in the same scale. 
Similar to the encoder, these network accept the concatenation of all the local latent variables and they generate the parameters of the same type of distribution.

In the decoder network, the ``context'' residual network first maps the concatenation of all the local latent variables and upsampled global latent variables to
a feature space. The output of this network is fed to a convolutional layer that generated parameters of a distribution on $\x_0$, which is subsampled 
$\x$ at scale $4\times4$. In the case of CIFAR10, the output of this layer correspond to the mixture of discretized logistic distribution \cite{salimans2017pixelcnn++} with 
10 mixtures. In the binary datasets, it is the parameters of a factorial Bernoulli distribution. The residual network input~$4\times4$ is applied to the
sample from this scale and its output is concatenated with the output of ``context''. The distribution on the next scale is formed similarly using another upsampling residual network. 
This process is repeated until we generate the image in the full scale. 

The residual blocks in our work consist of two convolutional layers with an skip connection. Resizing the dimensions is always handled in the first convolutional layer.
Downsampling is done using stride of 2 and upsampling is implemented using transposed convolution. The squeeze and excitation unit is applied with the reduction ratio $r = 4$ \cite{hu2017squeeze}.

The SELU \cite{klambauer2017self} and ELU \cite{clevert2015fast} activation functions are used in all the fully connected and convolutional layers respectively. 
No batch normalization was used except in the input of the encoder, the output of ``down 1'', and ``down 2''. AdaMax \cite{kingma2014adam} is used for training all the models. The learning rate
is set to 0.001 and is decreased when the value of the variational bound on the validation set plateaus. The batch size is 100 for all the experiments. In all experiments, the $\beta$ smoothing parameter is set to $8$.

We use parallel tempering~\cite{hukushima1996exchange, iba2001extended} to approximate the partition function of the RBM which is required
to evaluate generative log-likelihoods. We use chains at an adaptive number of temperatures, and perform 100,000 sweeps over all variables to ensure that the $\log Z$ estimate is reliable.

\subsection{Conditional Decoder} \label{app:cond_decod}
The conditional decoder for a $16\times16$-pixel image is illustrated in Fig.~\ref{fig:multi-scale}.

\def\NumOfColumns{16}%
\def\Sequence{0, 1, 2, 3, 4, 5, 6, 7, 8, 9, 10, 11, 12, 13, 14, 15}%

\newcommand{\Size}{0.3cm}
\tikzset{Square0/.style={
    inner sep=0pt,
    text width=\Size, 
    minimum size=\Size,
    draw=black,
    fill=red!40,
    align=center,
    }
}

\tikzset{Square1/.style={
    inner sep=0pt,
    text width=\Size, 
    minimum size=\Size,
    draw=black,
    fill=green!40,
    align=center,
    }
}

\tikzset{Square2/.style={
    inner sep=0pt,
    text width=\Size, 
    minimum size=\Size,
    draw=black,
    fill=blue!30,
    align=center,
    }
}

\begin{figure}
\centering
\begin{tikzpicture}[draw=black, thick, x=\Size,y=\Size]
    \foreach \row in \Sequence{%
        \foreach \col in \Sequence {%
            \pgfmathtruncatemacro{\value}{\col+\NumOfColumns*(\row)}
            \def\NodeText{\pgfmathparse{int(1 - \intcalcMod{\row+1}{2} * \intcalcMod{\col+1}{2})}\pgfmathresult}
            \pgfmathsetmacro{\RowMod}{\intcalcMod{\row}{4}}
            \pgfmathsetmacro{\ColMod}{\intcalcMod{\col}{4}}
            \pgfmathsetmacro{\ColModd}{\intcalcMod{\col}{2}}
            \IfEq{\RowMod}{0}{
              \IfEq{\ColMod}{0}{
                \node [Square0] at ($(\col,-\row)-(0.5,0.5)$) {};
              }{
                \IfEq{\ColMod}{2}{
                  \node [Square1] at ($(\col,-\row)-(0.5,0.5)$) {};
                }{
                  \node [Square2] at ($(\col,-\row)-(0.5,0.5)$) {};
                }
              }
            }{
              \IfEq{\RowMod}{2}{
                \IfEq{\ColModd}{0}{
                  \node [Square1] at ($(\col,-\row)-(0.5,0.5)$) {};
                }{
                  \node [Square2] at ($(\col,-\row)-(0.5,0.5)$) {};
                }
              }{
              \node [Square2] at ($(\col,-\row)-(0.5,0.5)$) {};
              }
            }
        }
    }
\end{tikzpicture}
\caption{In the conditional decoder we decompose an image into several scales. The generative process starts from the subset of pixels, $\color{red!80}\x_0$, at the lowest scale ($4\times4$). Conditioned on $\color{red!80}\x_0$, we generate the $8\times 8$ subset, $\color{green!100}\x_1$, at the next larger scale. Conditioned on $\color{red!80}\x_0$ and $\color{green!100}\x_1$, we generate the $16\times16$ subset, $\color{blue!80}\x_2$, at the next larger scale. This process is repeated until all pixels are covered. In this figure, pixels in $\color{red!80}\x_0$/$\color{green!100}\x_1$/$\color{blue!80}\x_2$ are indicated by their corresponding color. In order to have a consistent probabilistic model, given $\x_{<i}$, we only generate the remaining pixels, $\x_i$, at scale $i$.}
\label{fig:multi-scale}
\end{figure}

\section{Balancing the KL Term} \label{app:KLBalance}

With many layers of latent variable, the VAEs tend to disable many stochastic variables \cite{bowman2016generating, sonderby2016ladder}. Common mitigations
to this problem include KL annealing \cite{sonderby2016ladder}, free bits \cite{kingma2016improved}, and soft free bit \cite{chen2016variational}. 

In our experiments, we observe that annealing the KL term is more effective in maintaining active latent
variables than the free bits method. Nevertheless, at the end of training, the units tend to be
disabled unevenly across different groups. Some are completely inactive while other groups have many active variables. To address this, 
we modify the VAE objective function to
\begin{equation*}
\E_{q(\z|\x)}\left[ \log p(\x|\z)  \right] - \gamma \sum_i \alpha_i \KL(q(\z_{i}|\x) || p(\z_{i})).
\end{equation*}
$\gamma$ is annealed from zero to one during training, and $\alpha_i$ is introduced to balance the KL term across variable groups. 
As in soft free bits,
we reduce $\alpha_i$ if the $i^{th}$ group has a lower KL value in comparison to other groups and increase it if the KL value is higher for the group. 
In each parameter update $\alpha_i$ is determined as
\begin{equation*}
\alpha_i = \frac{N \hat{\alpha}_i}{\sum_j \hat{\alpha}_j} \ \ \text{where} \ \ \hat{\alpha}_i =  \E_{\x \sim \mathcal{M}} [\KL(q(\z_{i}|\x) || p(\z_{i}))] +\epsilon.
\end{equation*}
$N$ is the number of latent groups, $\mathcal{M}$ is the current mini-batch, and $\epsilon=0.1$ is a small value that softens the coefficients for very small values of $KL$. 
In this way, a group is penalized less in the KL term
if it has smaller a KL value, thereby encouraging the group to use more latent variables. We apply a stop gradient operation on $\alpha_i$ to prevent the AD from 
backpropagating through these coefficients. The $\alpha_i$ are included in the objective only while $\gamma$ is annealed. After $\gamma$  saturates at one, we set all $\alpha_i=1$ to allow the model to maximize the variational lower bound.  

\section{Additional Ablation Experiments}  \label{app:ablation}

In this section, we provide additional ablation experiments that target individual aspects of DVAE++. The test-set evaluations reported in this section do not use the binary model ($\beta=\infty$), but instead use the same $\beta$ that was used during training.
\begin{table} 
\centering
  \caption{The generative performance of DVAE++ improves with the number of local variable groups. 
  Performance is measured by test set log-likelihood in bits per dim.}
\begin{tabular}{ l|c|c|c|c|c}    
\# groups              &  8 & 12 & 16 & 20 & 24 \\
\hline
Bits per dim.          & 3.45 & 3.41 & 3.40 & 3.40 &  3.39 \\
\end{tabular} \label{tab:num_layer_expr}
\end{table}

\begin{table}
\centering
  \caption{The performance of DVAE++ improves with the number of global variable groups in the inference model 
  (i.e. $q(\z_i | \x, \bzeta_{<i})$). Performance is measured by test set log-likelihood in bits per dim.}
\begin{tabular}{ l|c|c|c}    
\# groups           & 1     & 2      & 4    \\
\hline
Bits per dim.       & 3.39 & 3.38  & 3.37 \\
\end{tabular} \label{tab:num_layer_global}
\end{table}

\setlength{\tabcolsep}{2pt}
\begin{table}
\caption{DVAE++ with and without conditional decoder} \label{tab:ab_cond}
\small
\centering
\begin{tabular}{l r r r r r}
\multirow{2}{*}{Baseline}  & MNIST & MNIST & OMNI- & Caltech- & CIFAR10 \\
  & (static) & (dynamic) & GLOT  & 101 &  \\
\hline
Conditional   & -79.37 & -78.62 & \textbf{-92.36} & \textbf{-81.85} & \textbf{3.37} \\
\hline
Unconditional & \textbf{-79.12} & \textbf{-78.47} & -92.94 & -82.40 & 3.91 \\
\hline
\end{tabular}
\end{table}

\begin{table}
\caption{DVAE++ is compared against baselines with either 
no global latent variables or no KL balancing coefficients.} \label{tab:ab_gl_kl}
\small
\centering
\begin{tabular}{l c c r}
Baseline  & MNIST & OMNIGLOT & CIFAR10 \\
  & (static) & & \\
\hline
DVAE++                   & -79.12 & \textbf{-92.36} & \textbf{3.37}\\
\hline
DVAE++ w/o global latent & \textbf{-78.96} & -92.60 & 3.41 \\
\hline
DVAE++ w/o kl balancing  & -79.72 & -92.74 &  3.42 \\
\hline
\end{tabular}
\end{table}

\subsection{Hierarchical Models Help}
As expected, we observe that increasing the number of local variable groups (the number of hierarchical levels) 
improves the performance of the generative model. 
Table~\ref{tab:num_layer_expr} summarizes the performance of the
DVAE++ for CIFAR10 as the hierarchy of continuous local variables is increased. 
Similarly, when global latent variables are modeled 
by an RBM, dependencies between discrete latent variables can develop. Modeling of these dependencies can 
require a deeper hierarchical inference model.
Table~\ref{tab:num_layer_global} summarizes the performance of
DVAE++ on CIFAR10. In this experiment the number of local groups is fixed to 16. The RBM consists of 128 binary variables and the number of hierarchical levels in the inference model is varied from 1 to 4. Deeper inference models generate high log-likelihoods (low bits per dimmension).

\subsection{Conditional vs. Unconditional Decoder}
In Table~\ref{tab:ab_cond}, the performance of DVAE++ with and without conditional decoder is reported. Multi-scale conditional decoders improve generative performance in all the datasets but MNIST.

\subsection{Global Latent Variables and KL Balancing}
Table~\ref{tab:ab_gl_kl} compares the performance of DVAE++ with 
global latent variables trained with KL balancing with two baselines on three datasets. 
In the first baseline, the global latent variables are completely removed from the model. 
In the second baseline, the KL balancing coefficient 
($\alpha_i$ in Sec.~\ref{app:KLBalance}) is removed from original DVAE++.
Removing either the global latent variables or the balancing coefficients typically decreases the
performance of our generative model. However, on binarized MNIST, DVAE++ without a global prior attains 
a new state-of-the-art result at -78.96 nats.

\end{document}